\documentclass[amsmath,amssymb,aps,pre,twocolumn]{revtex4-2}
\usepackage{graphicx}
\usepackage{dcolumn}
\usepackage{bm}
\usepackage{hyperref}
\usepackage{subfig}
\usepackage{xcolor}
\usepackage{CJKutf8}

\begin{document}

\title{\textbf{Average shortest-path length in word-adjacency networks: Chinese versus English}}

\author{Jakub Dec$^1$}
\author{Michał Dolina$^1$}
\author{Stanisław Drożdż$^{1,2}$}\email{Contact author: stanislaw.drozdz@ifj.edu.pl}
\author{Jarosław Kwapień$^2$}
\author{Jin Liu$^3$}
\author{Tomasz Stanisz$^2$}

\affiliation{$^1$ Faculty of Computer Science and Telecommunications, Cracow University of Technology, ul. Warszawska 25, 31-155 Krak\'ow, Poland}
\affiliation{$^2$ Complex Systems Theory Department, Institute of Nuclear Physics, Polish Academy of Sciences, ul. Radzikowskiego 152, 31-342 Krak\'ow, Poland}
\affiliation{$^3$ School of Modern Languages, Georgia Institute of Technology, Swann Building, 613 Cherry Street NW, Atlanta GA 30332-0375, USA}

\date{\today}

\begin{abstract}
Complex networks provide powerful tools for analyzing and understanding the intricate structures present in various systems, including natural language. Here, we analyze topology of growing word-adjacency networks constructed from Chinese and English literary works written in different periods. Unconventionally, instead of considering dictionary words only, we also include punctuation marks as if they were ordinary words. Our approach is based on two arguments: (1) punctuation carries genuine information related to emotional state, allows for logical grouping of content, provides a pause in reading, and facilitates understanding by avoiding ambiguity, and (2) our previous works have shown that punctuation marks behave like words in a Zipfian analysis and, if considered together with regular words, can improve authorship attribution in stylometric studies. We focus on a functional dependence of the average shortest path length $L(N)$ on a network size $N$ for different epochs and individual novels in their original language as well as for translations of selected novels into the other language. We approximate the empirical results with a growing network model and obtain satisfactory agreement between the two. We also observe that $L(N)$ behaves asymptotically similar for both languages if punctuation marks are included but becomes sizably larger for Chinese if punctuation marks are neglected.
\end{abstract}


\maketitle

\section{Introduction}
\label{sect::introduction}

Natural language, which serves as the primary medium of information in human societies, could not be a random mixture of elements and, thus, must contain strong internal correlations. This makes the choice of an appropriate representation - one that reveals these correlations and makes them accessible for analysis - crucial in the study of language structure. If we consider words as the basic functional units of language, such correlations are easily observed by examining word adjacency~\cite{KwapienJ-2012a,StaniszT-2024a}. As it is well known, some words cannot occur next to each other, while others frequently co-occur. This results from grammatical rules, context, and the style of the author of a given message. Similarly, differences in word usage frequency may also affect, which word pairs tend to appear together more or less frequently. Wherever relationships between the components of a larger system are considered, networks offer a convenient way to represent this system~\cite{MitchellM-2006a,CongJ-2014a,BattistonF-2020a}. In the context of natural language and, especially, written texts, words or phrases can be represented as nodes and edges between them can be based on various linguistic relationships such as co-occurrence, semantic similarity, or syntactic dependency. If one considers word co-occurrence networks, the nodes represent words and the edges represent their mutual adjacency if the words appear next to each other at least once within the analyzed text. Such networks, which fall into the category of evolving networks~\cite{DorogovtsevSN-2002a}, reveal a hierarchical, scale-free, disassortative structure with hubs and peripheral nodes, which is characterized by the small-world property and varying degree of network complexity~\cite{FerreriCanchoR-2001a,FosterJG-2010a,AmancioDR-2012a,WachsLopesGA-2016a}.

This structure of word-adjacency networks encodes both the presence of correlations (through linked nodes) and the absence or prohibition of certain word pairings (through missing edges). By assigning weights to edges in proportion to the frequency with which word pairs co-occur, one can quantify the strength of these correlations. When applied to different texts, which may vary by topic, authorship, or historical period, each with its own linguistic characteristics, the topology of the resulting networks can differ significantly. These structural variations can be exploited, for instance, in stylometric analysis aimed at author attribution or genre classification~\cite{SegarraS-2015a,AkimushkinC-2017a,StaniszT-2019a}. Different language groups and individual languages within a given group can also differ if presented in a network representation~\cite{LiuH-2013a} Of course, it is also possible to conduct analyses that account for higher-order correlations, taking into consideration not only word pairs but also triplets and larger groups of co-occurring words~\cite{MajhiS-2022a}.

In the classical approach to statistical linguistics, the fundamental elements of a text that carry meaningful information are words, defined as sequences of alphabetic characters separated by spaces~\cite{MillerGA-1957a}. A typical example of this approach is frequency analysis, in which words occurring in a language sample are counted, and the relationships between their frequencies are examined, often described by Zipf's law~\cite{ZipfGK-1932a}. This approach neglects everything in a text that falls outside of the words, such as, e.g., punctuation marks. Yet their role in written language is difficult to overstate: they serve to disambiguate meaning, highlight segments that form functional units, convey emotional tone, and provide the reader with cues for pauses or breathing. From this perspective, punctuation marks carry genuine information and omitting them may result in a diminished communicative nuance. The need to treat punctuation marks on equal footing with regular words becomes particularly striking within data analysis, where this approach leads to a better approximation of the rank-frequency relation by power-law distributions or to an improved text categorization~\cite{KuligA-2017a,StaniszT-2019a}. Punctuation marks play also an important role in building up long-range correlations in texts in a way that is similar to that played by words~\cite{StaniszT-2023a,StaniszT-2024a,StaniszT-2024b,DecJ-2024a,DecJ-2025a}.

Different languages can use different punctuation patterns, which can imply both distinct symbol sets and distinct conventions for their application. This is notably manifest between languages that use different writing systems. For example, when comparing the Western Indo-European languages, which use alphabetic scripts and possess a well-established set of punctuation conventions developed inherently over centuries, with Chinese, a language based on a logographic writing system with its own specific punctuation, on which the Western-style punctuation was imposed in the early 20th century and was normalized only half a century later. As a result, Chinese punctuation reflects a hybrid system, combining Western conventions with native elements.

In this study, a collection of texts written in Chinese over the past 150 years was analyzed. The texts were transformed into networks of word and punctuation mark co-occurrences, treating both types of elements equally. As a reference, additional networks were constructed using only words, with punctuation marks entirely excluded, which reflected the standard practice in conventional linguistic analyses. In addition, network representations of selected texts translated into English, as well as texts originally written in English translated into Chinese, were also examined. The main focus of the study was on a dynamical perspective, expressing how the network topology evolves as the network grows, which can be considered an equivalent of reading through the text. The topological property of the principal interest was the behavior of the average shortest path length (ASPL) as a function of the number of nodes in the network.

Although the Chinese language is spoken by approximately one billion people and is the most widely spoken first language in the world~\cite{ChenNF-2016}, its research in the complex network formalism is so far very limited~\cite{LiangW-2009a,LiJ-2012a,CrosleyG-2014a,JiangZ-2020a}. How does the inclusion of punctuation marks as functional linguistic elements affect the topology of word-adjacency networks in Chinese as compared to English texts, and what does this reveal about the relative structural and communicative role is the main research question addressed here.

\section{Methodology}
\label{sect::methodology}

\subsection{Punctuation in Chinese}

The Chinese punctuation possesses a distinct origin and developmental trajectory that its Western counterparts. In Chinese writing system, the fundamental unit is a character. Unlike Western languages, where words are generally constructed from sequences of letters, Chinese words consist of one or more characters. Characters may represent both morphemes (semantic units) and syllables, owing to the morphosyllabic nature of Chinese. Individual characters may serve either as independent words, often with multiple meanings, or be combined in n-grams to form compound words with varying interpretations depending on their context. This property led to a need for a punctuation system to improve reading comprehension and denote syntactic boundaries as early as around 900 BCE~\cite{GuanX-2002a}. Approximately 15 punctuation symbols, including one resembling the modern comma (","), were in use by the Han Dynasty (202 BCE -- 220 AD). The idea of sentence spacing began to take shape around the same period~\cite{GuanX-2002a}. The system become well developed by the late imperial period with a set of around fifty or sixty marks in active use, including the pause mark \begin{CJK}{UTF8}{gbsn}"、"\end{CJK}, the circle (sentence-ending) mark \begin{CJK}{UTF8}{gbsn}"。"\end{CJK}, and the title mark \begin{CJK}{UTF8}{gbsn}"《》"\end{CJK}. However, the system still lacked equivalents for Western punctuation marks such as the exclamation mark "!" or the question mark "?".

Despite this expansion, traditional Chinese punctuation remained non-standardized, leading to considerable variation and ambiguity. Many traditional punctuation symbols were not embedded within the text lines but instead appeared in the margins or interlinear spaces and served as external annotations or commentary. Moreover, official Classical Chinese documents typically omitted punctuation altogether, leaving sentence segmentation to the reader's interpretation. The adoption of Western-style punctuation in the early twentieth century significantly influenced writing conventions. Chinese authors gradually incorporated the imported punctuation into their writing in the 1920s and 1930s, embedding these marks directly within the text body~\cite{MullaneyTS-2017a}.

Longstanding conventions persist, however. The historical absence of clear distinctions between commas and periods in Chinese writing may help explain why modern Chinese authors frequently construct run-on sentences or clauses marked solely by commas or semicolons. Since sentence segmentation in Chinese remains an open and debated problem in natural language processing and there is still no broadly accepted standard for defining sentence boundaries based on punctuation~\cite{LiuJ-2023a}, this study adopts the following characters as sentence terminators:\begin{CJK}{UTF8}{gbsn} 。\end{CJK}(ideographic full stop), \begin{CJK}{UTF8}{gbsn}！\end{CJK}(fullwidth exclamation mark),\begin{CJK}{UTF8}{gbsn} ？\end{CJK}(fullwidth question mark), \begin{CJK}{UTF8}{gbsn}…\end{CJK} (horizontal ellipsis), \begin{CJK}{UTF8}{gbsn}；\end{CJK}(fullwidth semicolon), and \textit{newline}, provided that the symbol sequence is not otherwise terminated by the aforementioned marks. Additionally, Chinese writers frequently use parentheses \begin{CJK}{UTF8}{gbsn}(【 】)\end{CJK} and various hyphens to mark pauses, quotations, or emphasis; these marks were excluded from the present punctuation set. Characters such as single-angle brackets \begin{CJK}{UTF8}{gbsn}(〈 〉)\end{CJK} and "Chinese quotation marks" in the form of corner brackets \begin{CJK}{UTF8}{gbsn}(「 」)\end{CJK} also appear in written Chinese. Correct usage and placement of all these punctuation marks play a crucial role in conveying sentence-level meaning and enabling coherent written communication as well as, in this case, a proper construction of the word-and-mark adjacency networks.

\subsection{Network modeling}

Construction of a network model that could recreate dynamics of natural language is not a straightforward task. First, one has to take into consideration the universal properties of language expressed by the Zipf-Mandelbrot and Heaps laws. The Zipf-Mandelbrot law~\cite{MandelbrotBB-1953a,MandelbrotBB-1954a} describes a relation between frequency of word occurrence and word rank within a language sample, which occurs to be a shifted power-law function of the type:
\begin{equation}
f(r) \sim \frac{1}{(r+\beta)^{\alpha}},
\label{eq::zipf.mandelbrot}
\end{equation}
where $\alpha$ is the Zipf exponent, which can vary between languages, literary styles, and text samples but, typically, $0.7 \leqslant \alpha \leqslant 1.3$~\cite{HaLQ-2003a,Ferrer-i-CanchoR-2005a,KwapienJ-2010a,GrabskaGradzinskaI-2012a,KoplenigA-2018a,DolinaM-2025a,LiYanfang-2025a} and $\beta$ denotes a shift parameter which typically falls into the interval $0.4 < \beta < 4$ for the Indo-European languages~\cite{KoplenigA-2018a}. It can be derived by solving a problem of communication optimization~\cite{MandelbrotBB-1953a,MandelbrotBB-1954a}. Often, this law is approximated by putting $\beta=0$, which then becomes its original Zipf's formulation~\cite{ZipfGK-1932a,ZipfGK-1949a}. Another important relation is the Heaps law that describes a functional dependence of the vocabulary size of a text sample on the sample's length measured in words~\cite{HerdanG-1960a,HeapsHS-1978a}. This relation is also a power law:
\begin{equation}
N(\tau) \sim \tau^{\delta}
\label{eq::heaps}
\end{equation}
with typical values of the exponent $\delta$ being in the range $0.4 < \delta < 1$~\cite{BochkarevVV-2014a}. This relation means that with increasing text sample size, the rate of occurrence of new words decreases, and previously used words tend to be repeated more frequently.

From a word-adjacency network perspective, the power-law character of the word-frequency distributions may suggest that the dynamics of network growth can be captured by some variation of a preferential attachment model~\cite{BarabasiAL-1999a}. However, a sub-linear dependence expressed by Eq.~(\ref{eq::heaps}) makes a linear-growth model inadequate and instead shifts the focus toward a model of accelerated growth. The immediate candidate is the Dorogovtsev-Mendes model~\cite{DorogovtsevSN-2000a,DorogovtsevSN-2001b,DorogovtsevSN-2002b}, which has already been applied to model word-adjacency networks in both its original~\cite{DorogovtsevSN-2001a} and modified versions~\cite{MasucciAP-2006a,MarkosovaM-2008a,KuligA-2015a}. Although this model is able to reproduce some key network topological properties of the empirical data like the scale-free node-degree probability mass function, small-world property, negative assortativity coefficient, and scale-free dependence of local clustering coefficient on node degree, the average shortest path length, defined as
\begin{equation}
L(N) = \frac{1}{N(N-1)} \sum_{i,j} d(i,j),
\label{eq::aspl}
\end{equation}
where $d(i,j)$ is the shortest path between nodes $i$ and $j$, remains a more difficult property to be modeled properly. In part, this stems from a fact that a word-adjacency network whose growth is driven by the progressive addition of words to the text cannot be represented by a random network model in which new nodes and edges can be added to the existing network anywhere if only some predefined probability distribution is applied. In a real network the new nodes and intra-network edges are added always to the node that was connected by an edge in the preceding step. This property introduces memory that must be taken into account when a model is designed~\cite{MasucciAP-2006a,KuligA-2015a}.

When the text begins to be generated, in the initial phase, each step of the procedure introduces a new word that has not yet appeared before. During this phase, the network takes the form of a chain, with new nodes being added to its end. If the probability of using a new word is described by a function $p(\tau)$, where $\tau$ denotes the length of the text written so far, then $p(\tau) \approx 1$ in this initial phase. At some point, one of the most frequently used words in the language (such as a pronoun or article) occurs for the second time -- the network ceases to be a chain at that moment and the first cycle emerges. From this point forward, network evolution involves the decreasing probability of a new node addition and the increasingly frequent addition of edges between existing nodes. In models based on the accelerated growth paradigm, the probability evolves according to a power-law function $p(\tau) = p_0 \tau^{\delta-1}$, where $\delta < 1$. This guarantees that the Heaps law (Eq.~(\ref{eq::heaps})) is obeyed. If a new node is added, it must be connected to the node that was connected by an edge in the preceding step. Moreover, a new edge is added with probability $1-p(\tau)$ by connecting the node that was connected in the previous step with a node $i$ selected from the existing ones according to a time-dependent probability distribution $\pi(i,t)$, where $t$ is the network growth step such that $N(t) = t$. A particular form of this probability is model-dependent; one of the possible choices is
\begin{equation}
\pi(k_i,t) \sim k_i^{\xi(t)}, \quad \xi(t) = 1 - t^{-\eta},
\label{eq::accelerated.growth.edge.probability}
\end{equation}
where $k_i$ stands for degree of the node $i$, $c_1 > 0$, and $\eta > 0$. In this case, the network grows according to a sublinear preferential attachment, which becomes linear asymptotically, when the network is large and mature~\cite{KuligA-2015a}. Another possible choice of $\pi(i,t)$ is a mixture of a global and local preferential attachment~\cite{MasucciAP-2006a} or a preferential attachment with edge rewiring~\cite{MarkosovaM-2008a}.

An empirical form of the average shortest path length $L(N)$ shown in Figs.~\ref{fig::chinese.corpora.1}--\ref{fig::translation.matrix} consists of three regimes. Initially, it increases linearly as $L_{\rm chain}(N) = (N+1)/3$, which corresponds to the chain-formation phase described above. Then, this increase is gradually suppressed and $L(N)$ passes through its maximum value for $10 < N < 100$, which typically satisfies $L_{\rm max} < 10$, and starts decreasing if $N$ increases further. In the remaining third phase, $L(N)$ reveals an asymptotic decline as $N \to \infty$. It should be noted that the decreasing $L(N)$ is rather uncommon for networks without accelerated growth as even the ultrasmall-world networks develop an increasing average shortest path length~\cite{CohenR-2003a,DereichS-2012a}. This behavior can best be approximated by a model combining the accelerated growth with a nonlinear preferential attachment given by Eq.~(\ref{eq::accelerated.growth.edge.probability}) and a linear chain growth~\cite{KuligA-2015a}.

In our earlier work, it was shown that, if the network is large enough, its average shortest path length evolves similarly to the one for a random graph with average node degree $\langle k \rangle$. $L(N)$ can thus be approximated by using the following formula~\cite{KuligA-2015a}:
\begin{equation}
L_{\rm rand}(N) \sim \frac{\ln N}{\ln\frac{c_0}{\alpha+1} + \alpha \ln N},
\label{eq::aspl.large.network}
\end{equation}
where $c_0$ and $\alpha$ play the role of free parameters that can be fitted to an empirical function $L(N)$. In the thermodynamical limit, it assumes
\begin{equation}
\lim_{N \to \infty} L_{\rm rand}(N) = \frac{1}{\alpha}.
\label{eq::aspl.asymptotic}
\end{equation}
Here, $\alpha$ is a power-law index of the accelerated growth describing how many edges are added among the existing nodes for each newly added node: $c(t) = c_0 t^{\alpha}$~\cite{DorogovtsevSN-2000a,KuligA-2015a}.

In order to approximate the evolution of empirical $L(N)$ in the whole range of available values of $N$, we apply the sigmoid-based interpolation between the regimes for small and large $N$:
\begin{equation}
L_{\rm fit}(N) = (1-S(N))L_{\rm chain}(N) + S(N) L_{\rm rand}(N),
\label{eq::interpolation}
\end{equation}
where
\begin{equation}
S(N) = \frac{1}{1+\left( \frac{N_0}{N}\right)^{\theta}}
\label{eq::sigmoid.interpolant}
\end{equation}
is the sigmoid interpolant. In the above formula, the parameters $N_0$ and $\theta$ define the regime-transition point and transition sharpness, respectively.

Computation of ASPL for a network with $N$ nodes is a time-consuming task if $N$ exceeds a few thousand. The fastest exact algorithm is the Breadth-First Search that was used in this study; its complexity is $\mathcal{O}(N(N+E))$, where $E$ is the number of edges. This algorithm works particularly fast for networks of small or moderate size (up to, say, $10^5$ nodes) since $N \leqslant \mathcal{O}(E) \leqslant N^2$ depending on edge density $E$.

\section{Data}
\label{sect::data}

We analyzed a set of 94 books written in Chinese, which represented different literary epochs: (1) the Late Qing Era (denoted LQ, from the mid-19th century to 1911, 18 novels), (2) the Republican Era (RE, 1912--1948, 11 novels), (3) the Maoist Era (ME, 1949--1978, 10 novels), and (4) the Contemporary China Era (CC, since 1979, 23 novels). Moreover, we distinguished books whose authors worked in (5) Hong Kong (HK, 8 novels) and (6) Taiwan (TW, 9 novels) and considered them as two separate subsets. Finally, (7) we collected a set of novels published on the Internet and put them into a separate category (IT, 15 novels), because their authors took advantage of the specific functionalities of this medium, which allows for publication of significantly longer texts than traditional print media, as well as for the use of a much richer array of tools, such as emoticons.

In the section devoted to translations between Chinese and English, our corpus comprised five titles (three of them, C1-C3, were also included in the respective subsets mentioned above): \\
(C1) \textit{The Drunkard}, authored by Liu Yi-Chang and first published in Chinese in 1962 as a serial in a Hong Kong evening paper. It is considered one of the first full-length stream-of-consciousness novels written in Chinese. The English translation of this novel was done by Charlotte Chun-Lam Yiu. \\
(C2) \textit{The Soul Mountain} published in 1990 by Gao Xingjian, the recipient of the 2000 Nobel Prize in Literature. This novel is renowned for its unique narrative style and profound exploration of identity, nature, and the human condition. The corresponding English translation was done by Mabel Lee. \\
(C3) \textit{The Sun Shines over the Sanggan River} was authored by Ding Ling, who wrote it in 1948, describing lives of peasants and the class struggles during the implementation of land reform in northern China. It was translated into English by W.J.F.~Jenner. \\
(E1) \textit{Alice's Adventures in Wonderland} written by Lewis Carroll in 1865 lists among the most popular works of literature for children and young adults worldwide, it is also one of the most extensively analyzed literary texts. It has been translated into 174 languages, including Chinese. Our analysis focuses on two independent Chinese translations: one by Zhao Yuanren (labelled as ``translation 1'' here) and another one by Ma Ainong (``translation 2''). \\
(E2) \textit{David Copperfield}, authored by Charles Dickens and first published as a newspaper serial in 1849-50, is a story of coming of age, hardships, love, and self-discovery of the main protagonist. The book was translated into Chinese by Lin Shu.

Practical procedures for lexical segmentation of Chinese texts are facilitated by existing computational libraries offering pre-implemented functions. The present study employs the Jieba library for this purpose~\cite{jieba}. After segmenting each text into individual words (English texts were not lemmatized), word-adjacency networks were constructed from the texts by connecting word pairs that appear in direct succession at least once in a text. The number of co-occurrences for identical word pairs was not considered; as a result, the constructed networks are binary. For each text, two different network representations were created: (1) a network consisting of words only, in which punctuation marks were neglected during construction, such that two words were treated as adjacent even when a punctuation mark appeared between them, and (2) a full network, in which words and punctuation marks (collectively referred to as tokens henceforth) were treated equivalently and both types of objects were represented by nodes.

\section{Results and discussion}
\label{sect::results}

\begin{figure*}[t]
\centering
\hrule\vspace{5pt}
\textbf{The Drunkard - Chinese} \\
{\includegraphics[trim={0cm 0 0 3.5cm}, clip, width = 0.33\linewidth]{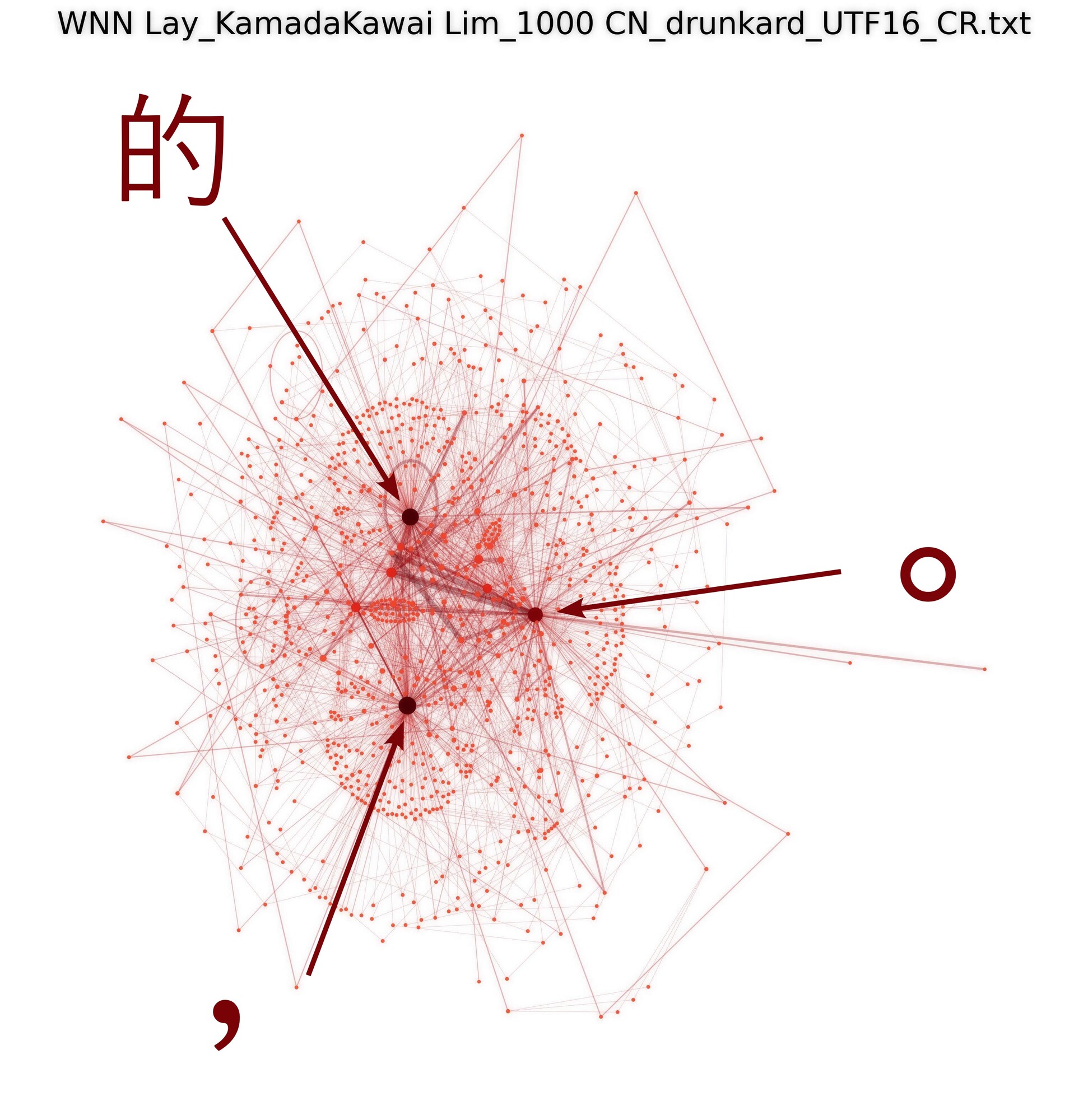}}
\hfill
{\includegraphics[width = 0.33\linewidth]{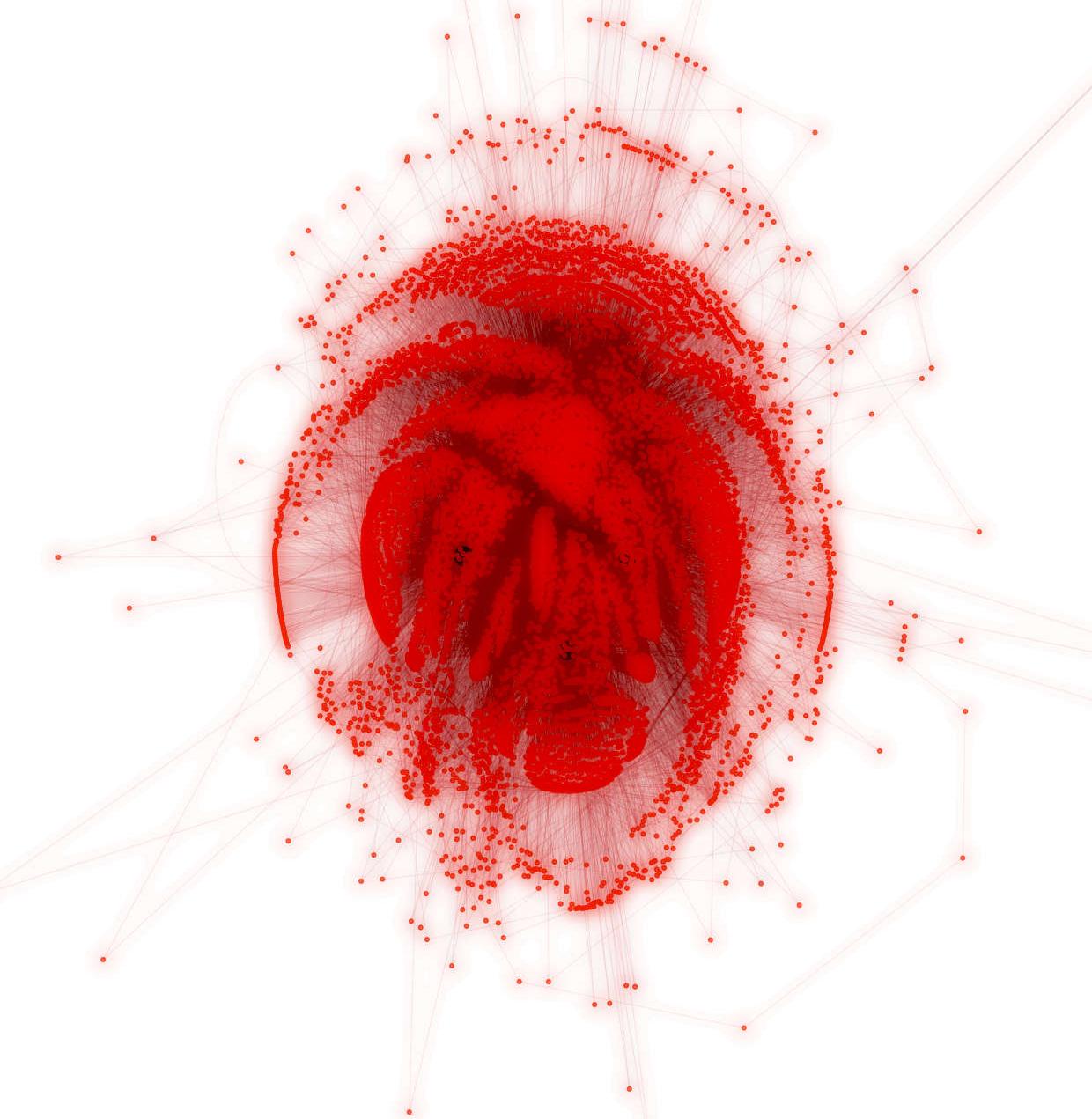}}
\hfill
{\includegraphics[trim={0cm 0 0 0.7cm}, clip, width = 0.32\linewidth]{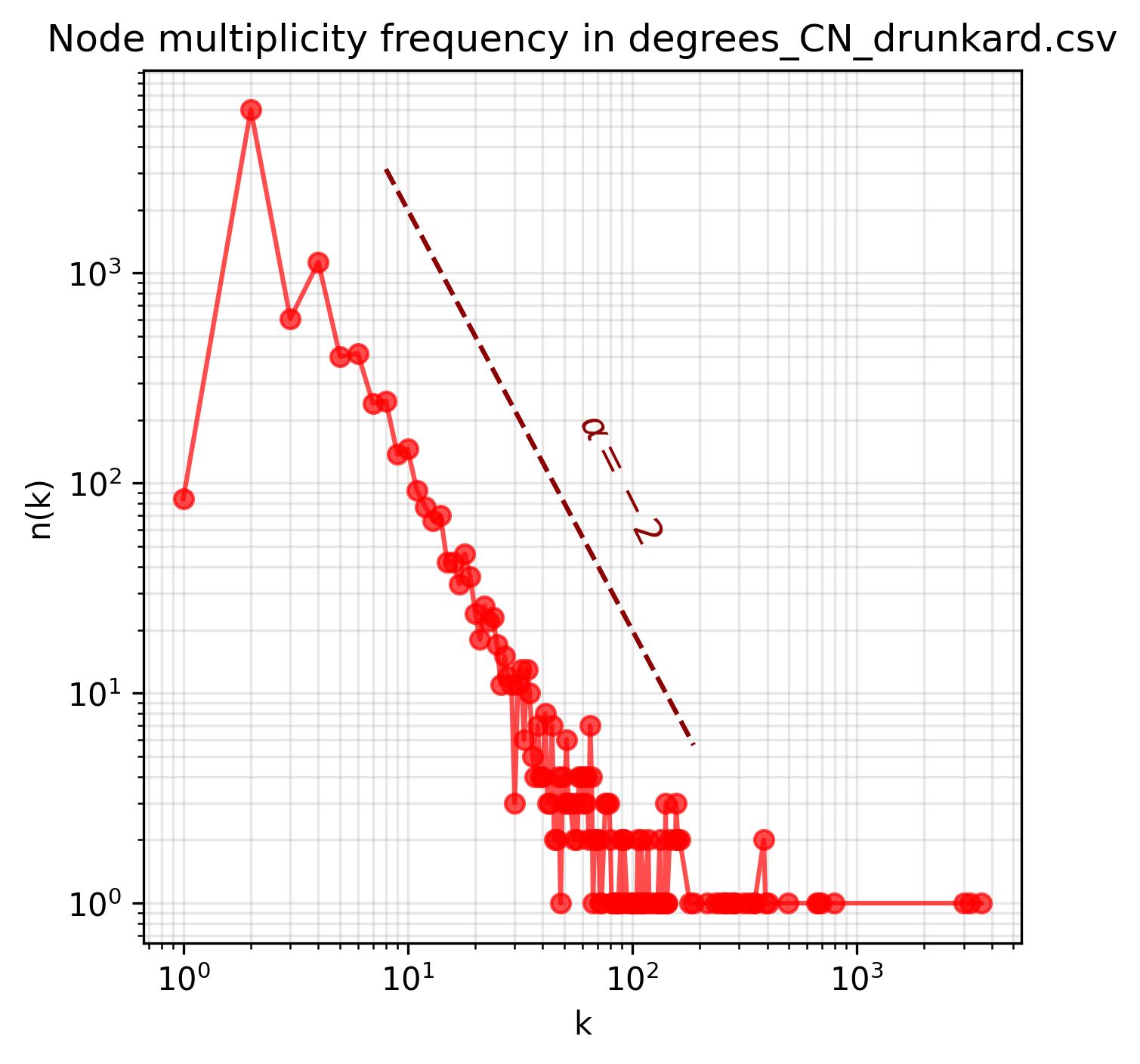}} \\

\textbf{The Drunkard - English} \\
{\includegraphics[trim={0cm 0 0 3.5cm}, clip, width = 0.33\linewidth]{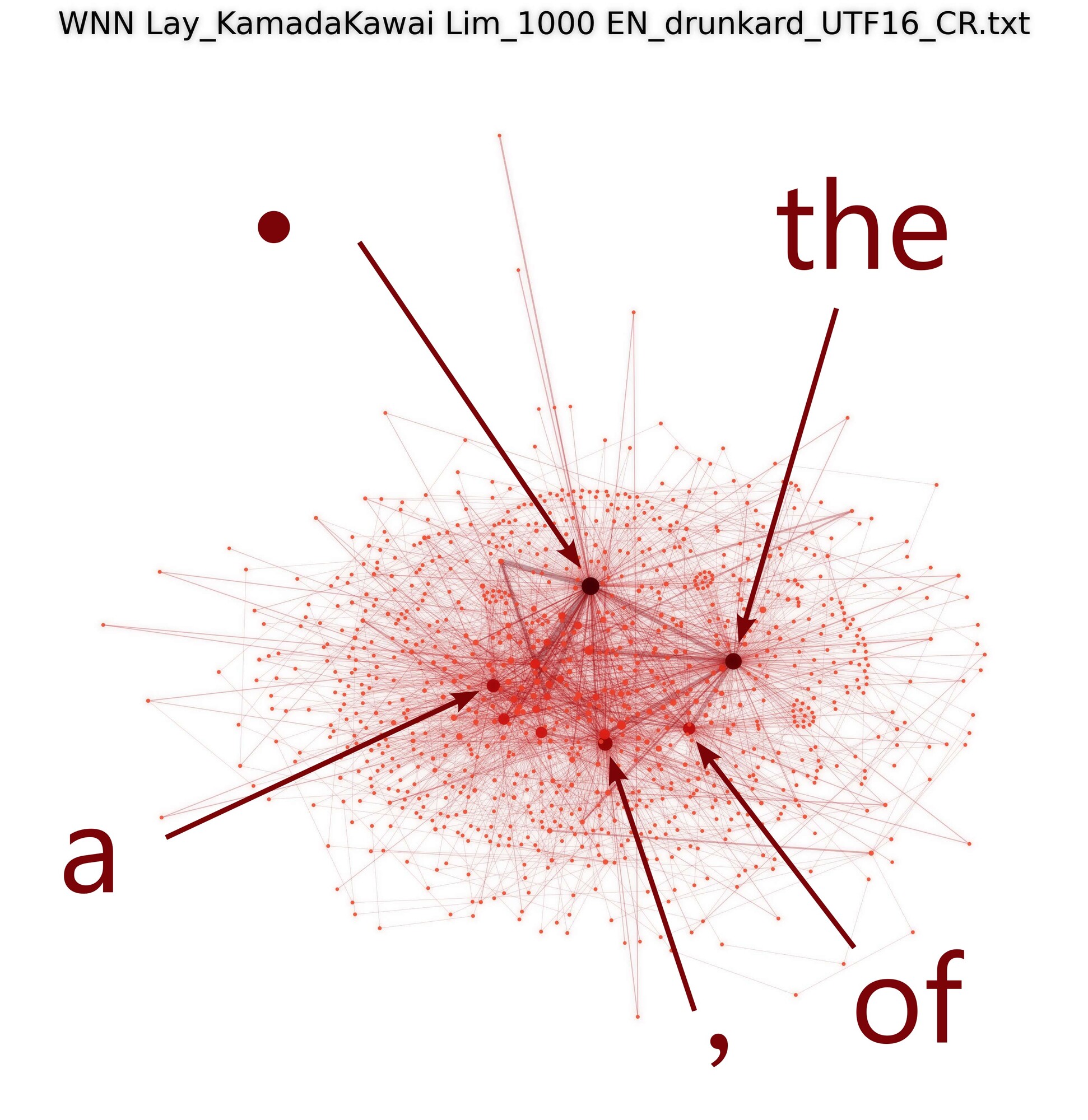}}
\hfill
{\includegraphics[width = 0.33\linewidth]{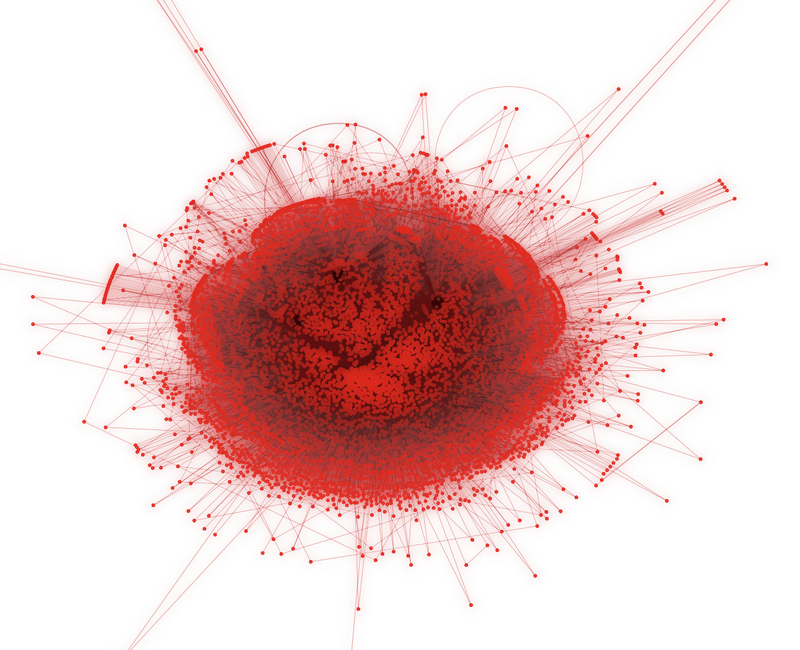}}
\hfill
{\includegraphics[trim={0cm 0 0 0.7cm}, clip, width = 0.32\linewidth]{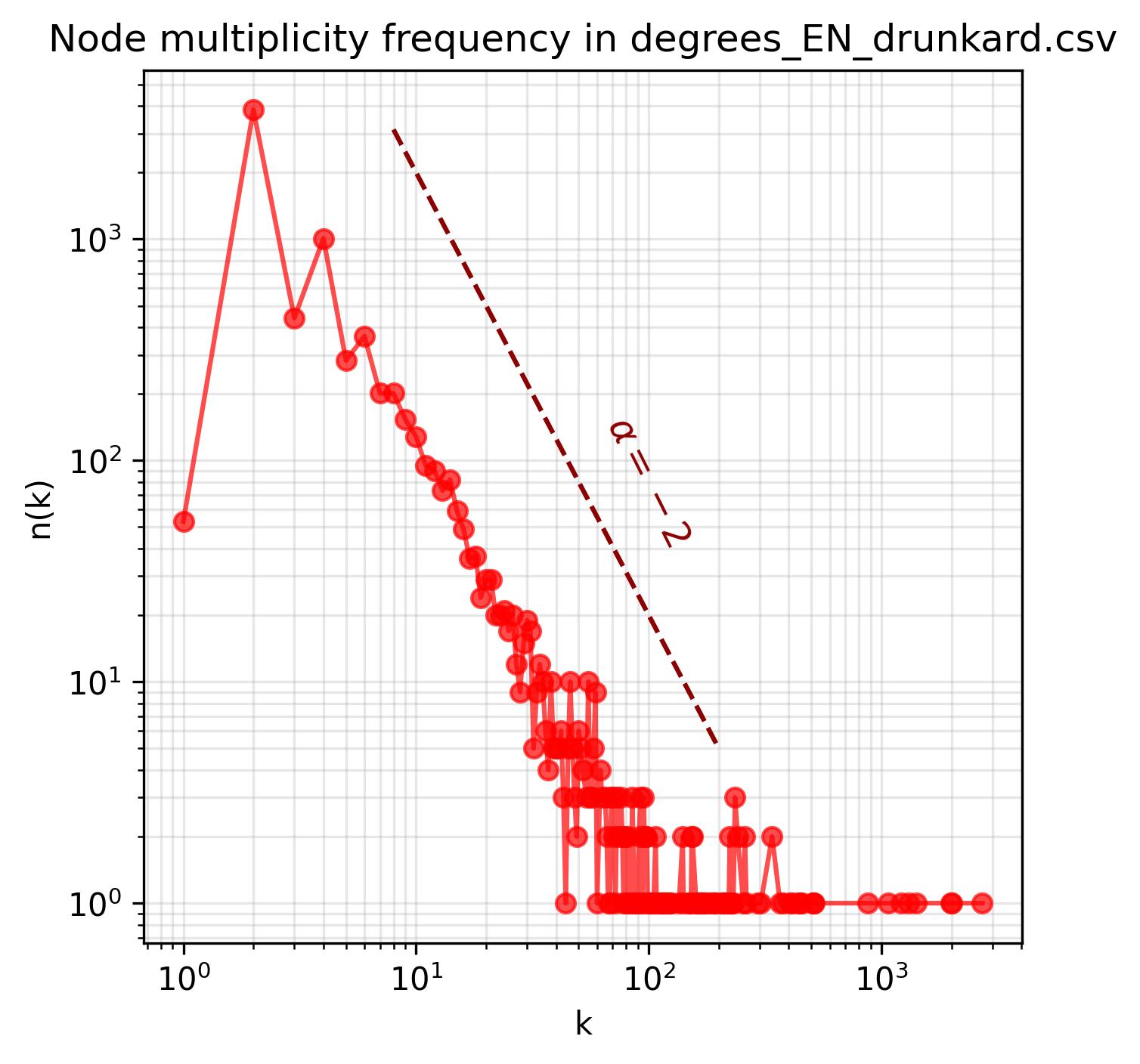}}

\parbox[c]{0.33\linewidth}{\centering First 1000 nodes} \hfill
\parbox[c]{0.33\linewidth}{\centering Entire book} \hfill 
\parbox[b]{0.3\linewidth}{\centering \# $k$-degree nodes}

\vspace{5pt}\hrule\vspace{5pt}
\caption{Word and punctuation-mark adjacency networks for \textit{The Drunkard} in Chinese and English. For each text, the left column corresponds to networks created from the first 1,000 unique words and punctuation marks, the middle column corresponds to networks created from the entire books, and the right column represents the node-degree distributions calculated for the entire books.}
\label{fig::drunkardnet}
\vspace{-0.4cm}
\end{figure*}

\begin{figure*}[t]
\centering
\hrule\vspace{5pt}
\textbf{Soul Mountain - Chinese} \\
{\includegraphics[trim={0cm 0 0 3.5cm}, clip, width = 0.33\linewidth]{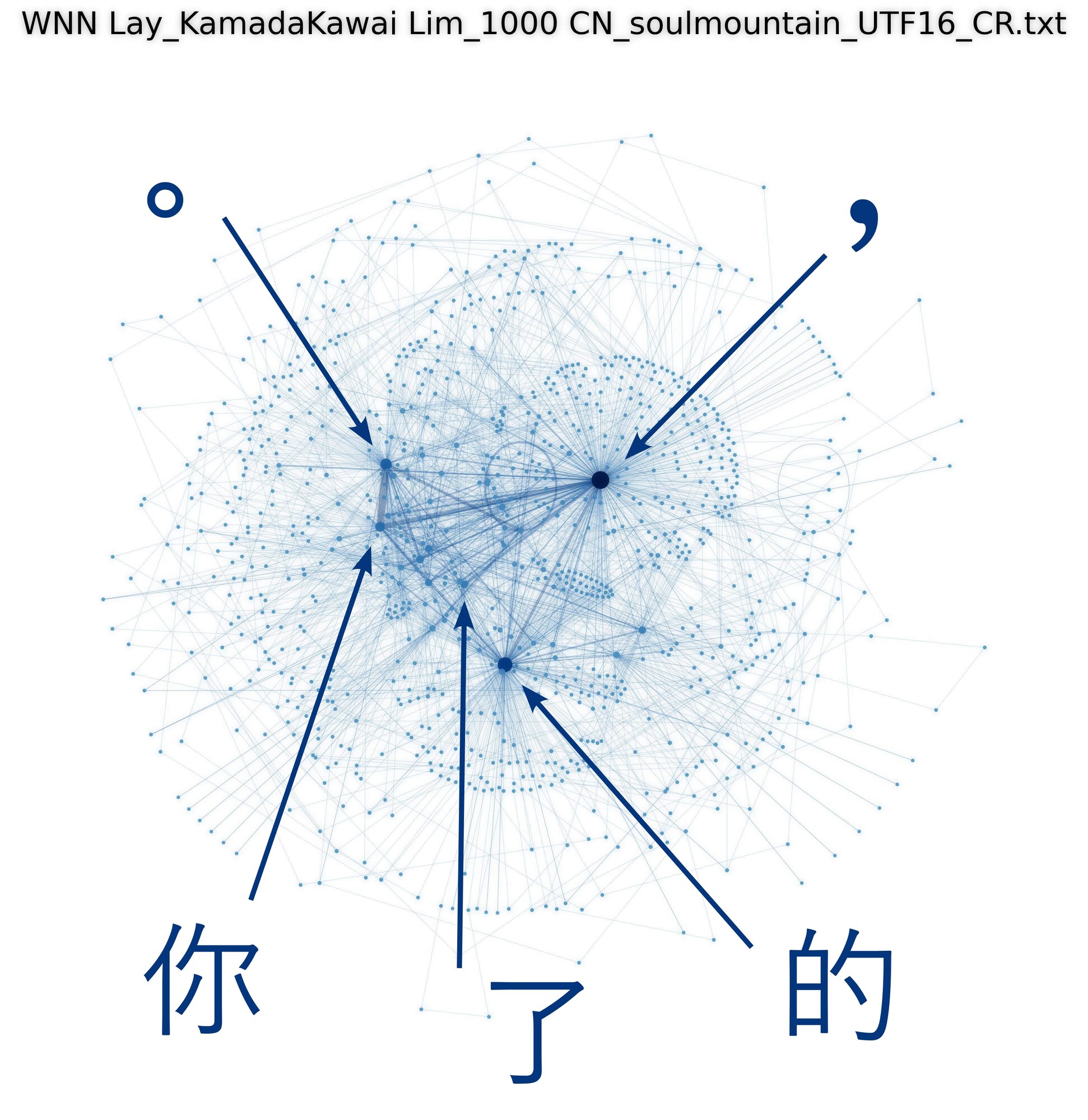}}
\hfill
{\includegraphics[trim={0cm 0 0 1.1cm}, clip, width = 0.33\linewidth]{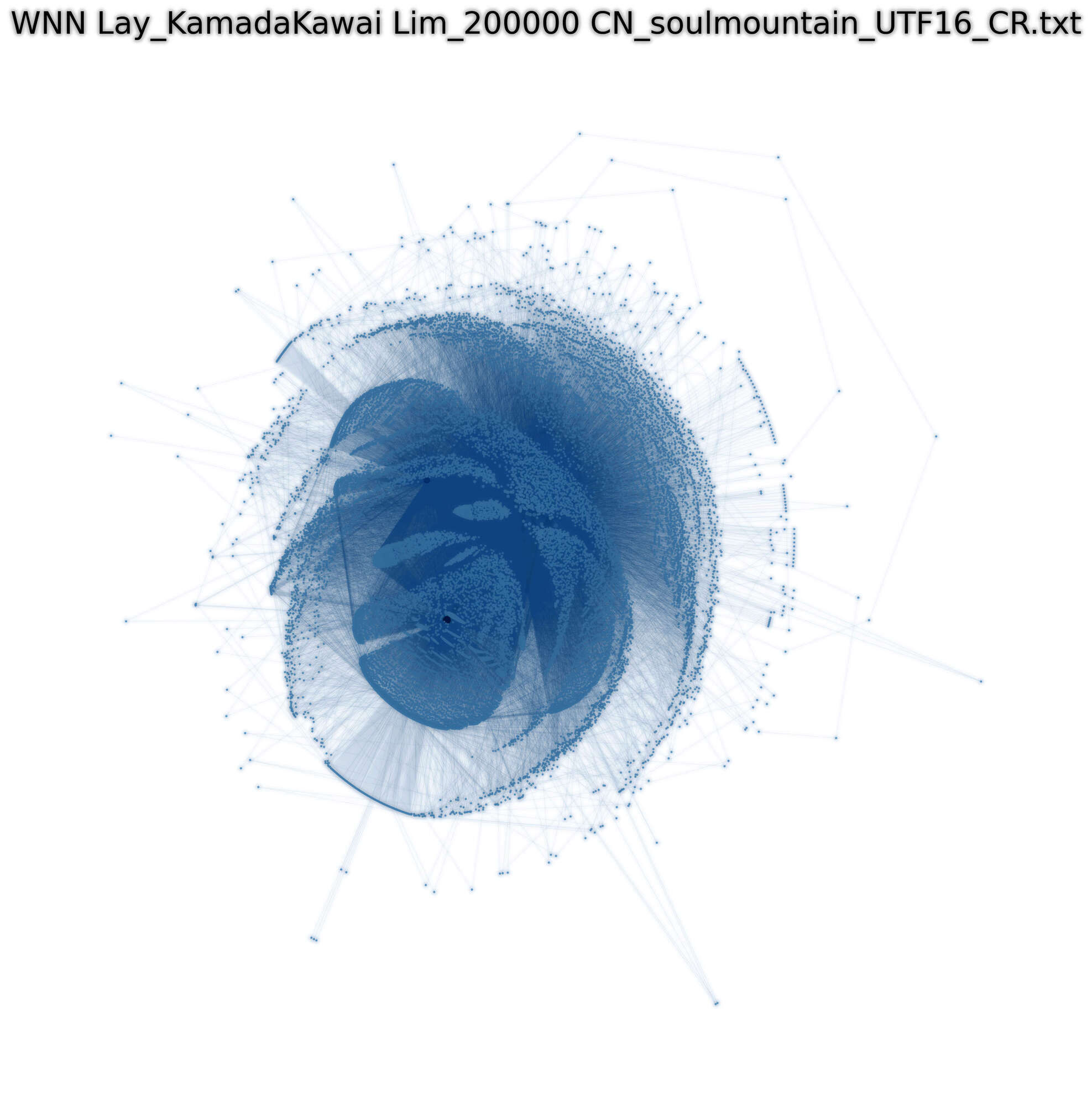}}
\hfill
{\includegraphics[trim={0cm 0 0 0.7cm}, clip, width = 0.32\linewidth]{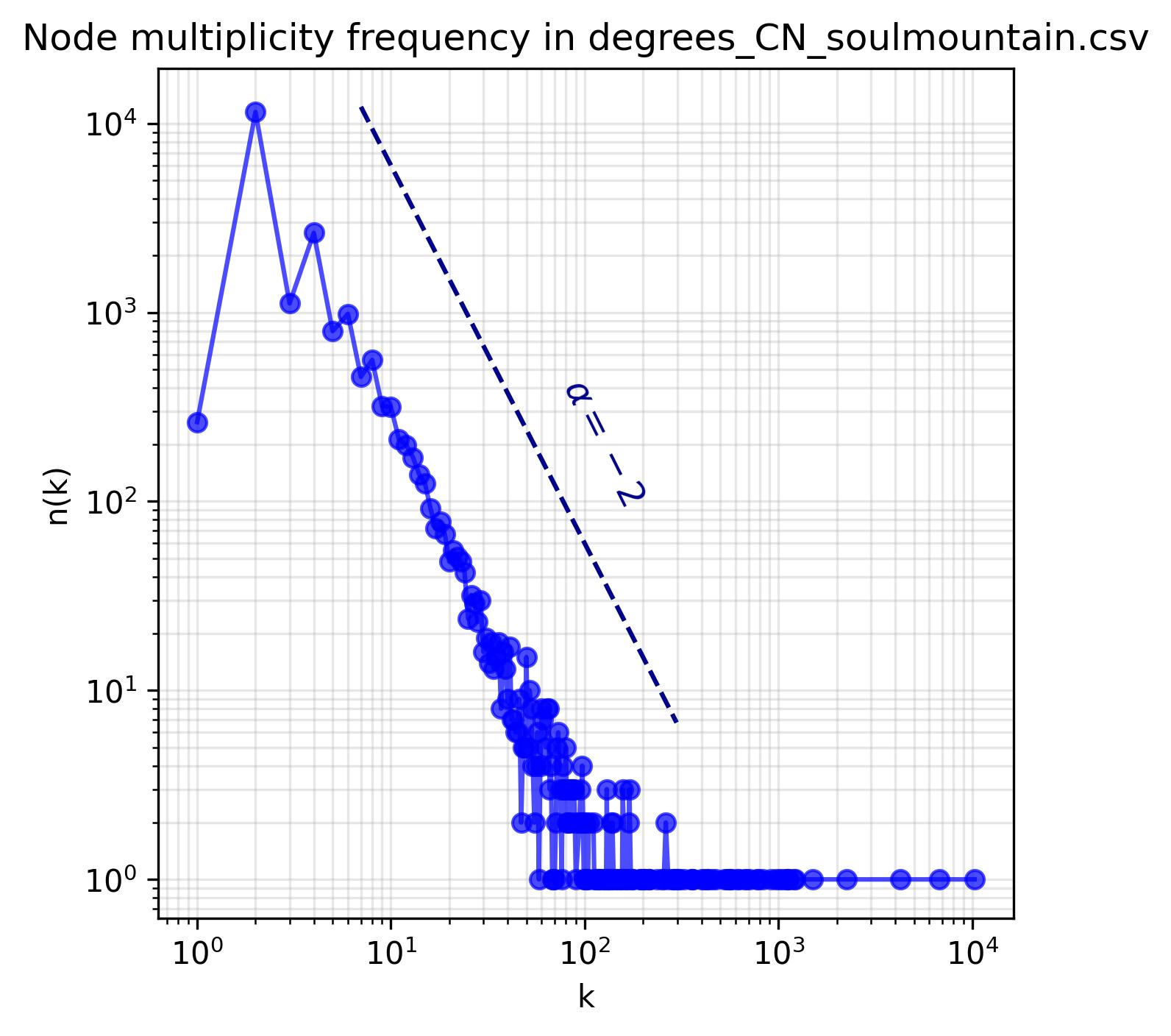}} \\

\textbf{Soul Mountain - English} \\
{\includegraphics[trim={0cm 0 0 3.5cm}, clip, width = 0.33\linewidth]{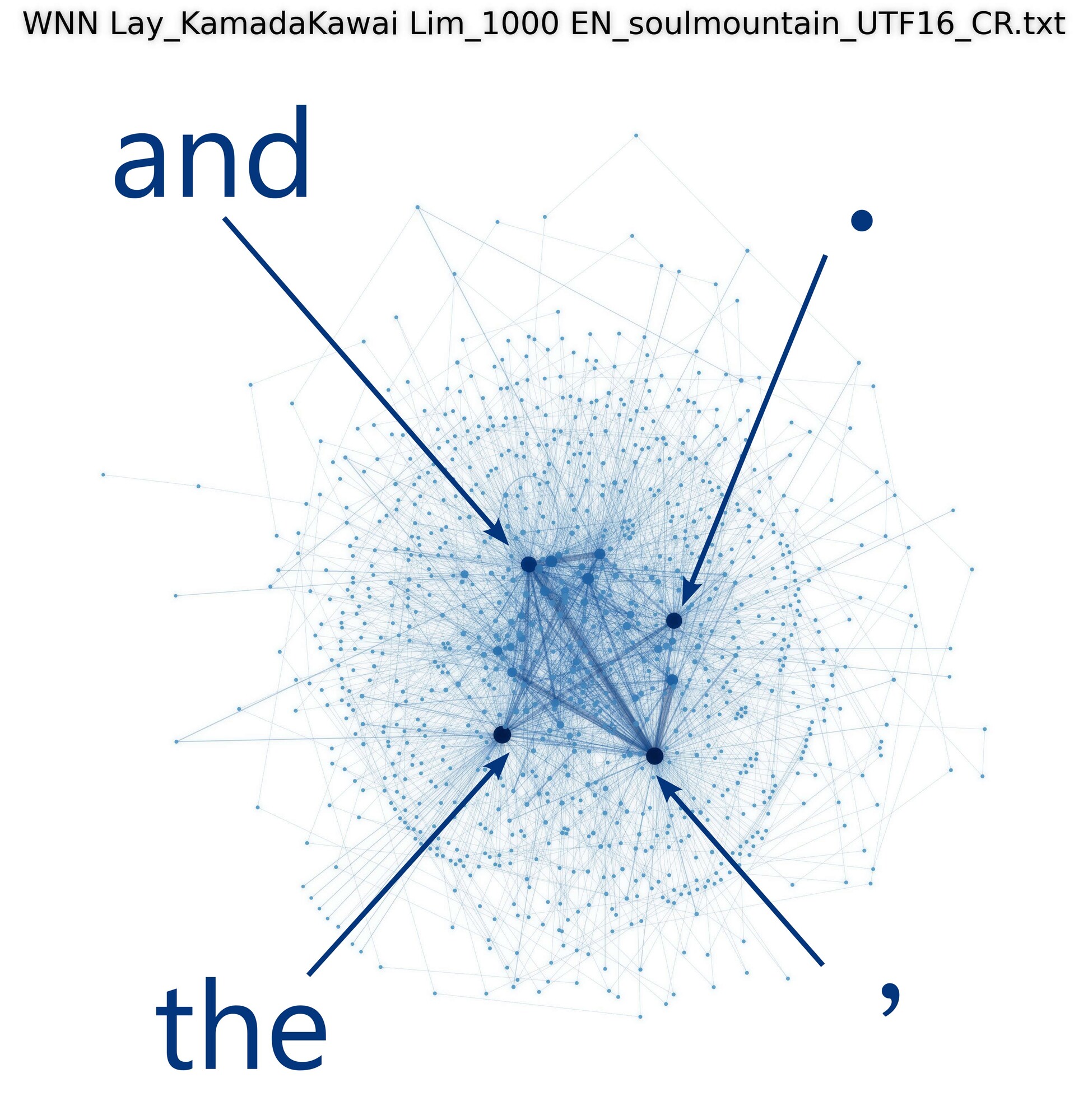}}
\hfill
{\includegraphics[width = 0.33\linewidth]{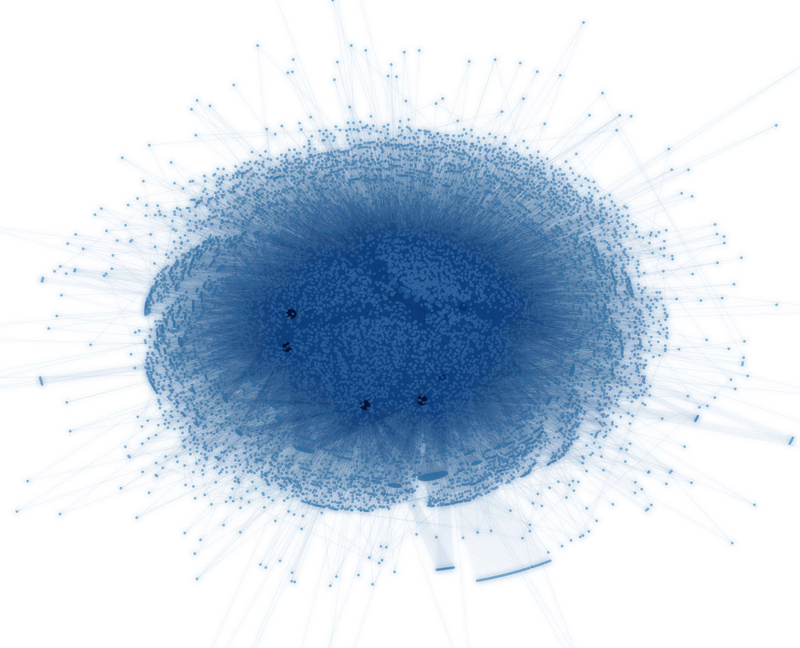}}
\hfill
{\includegraphics[trim={0cm 0 0 0.7cm}, clip, width = 0.32\linewidth]{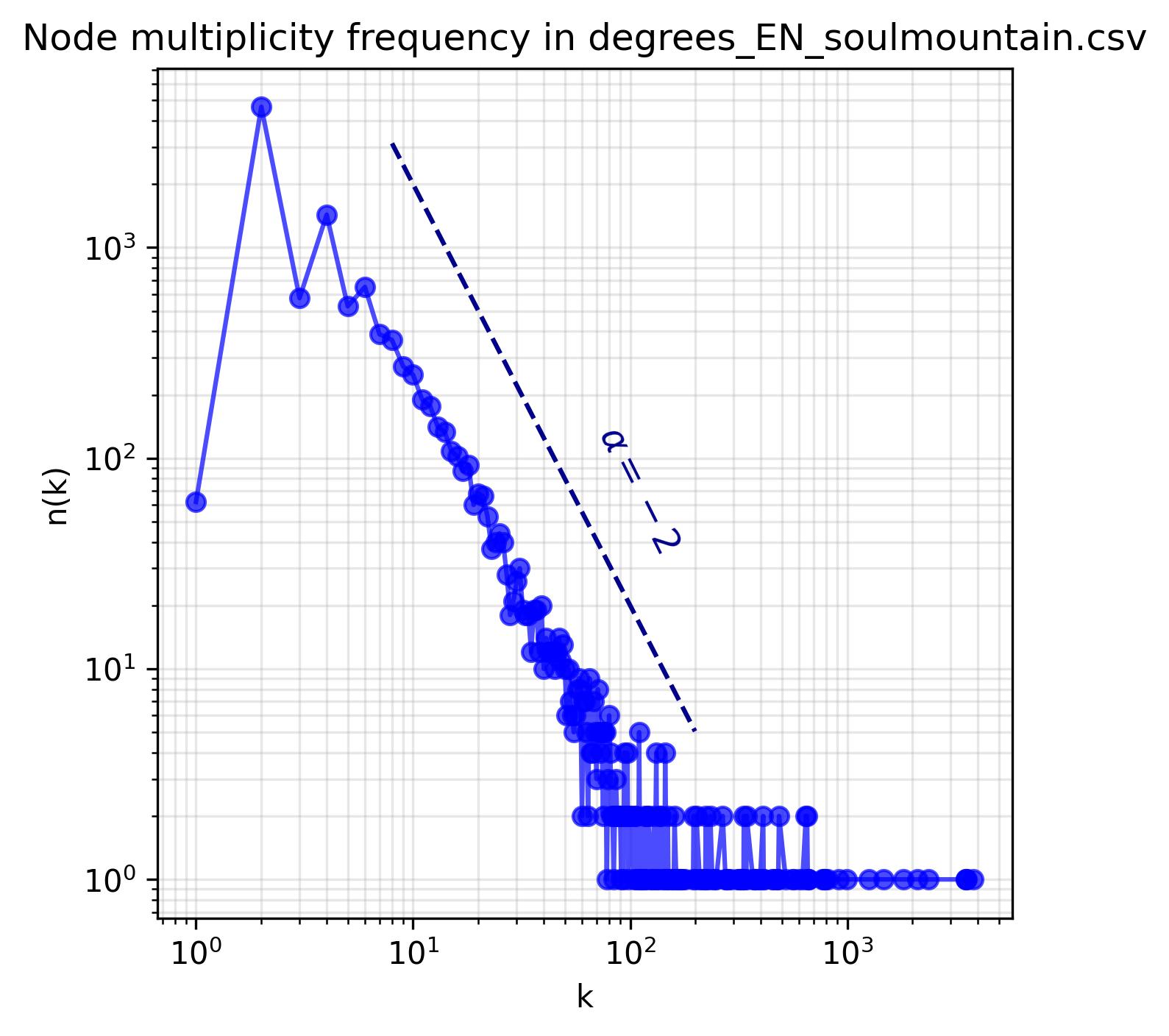}}

\parbox[c]{0.33\linewidth}{\centering First 1000 nodes} \hfill
\parbox[c]{0.33\linewidth}{\centering Entire book} \hfill 
\parbox[b]{0.3\linewidth}{\centering \# $k$-degree nodes}

\vspace{5pt}\hrule\vspace{5pt}
\caption{Word and punctuation-mark adjacency networks for \textit{Soul Mountain} in Chinese and English. For each text, the left column corresponds to networks created from the first 1,000 unique words and punctuation marks, the middle column corresponds to networks created from the entire books, and the right column represents the node-degree distributions calculated for the entire books.}
\label{fig::soulmountainnet}
\vspace{-0.4cm}
\end{figure*}

\begin{figure*}[t]
\centering
\hrule\vspace{5pt}
\textbf{The Sun Shines over the Sanggan River - Chinese} \\
{\includegraphics[trim={0cm 0 0 3.5cm}, clip, width = 0.33\linewidth]{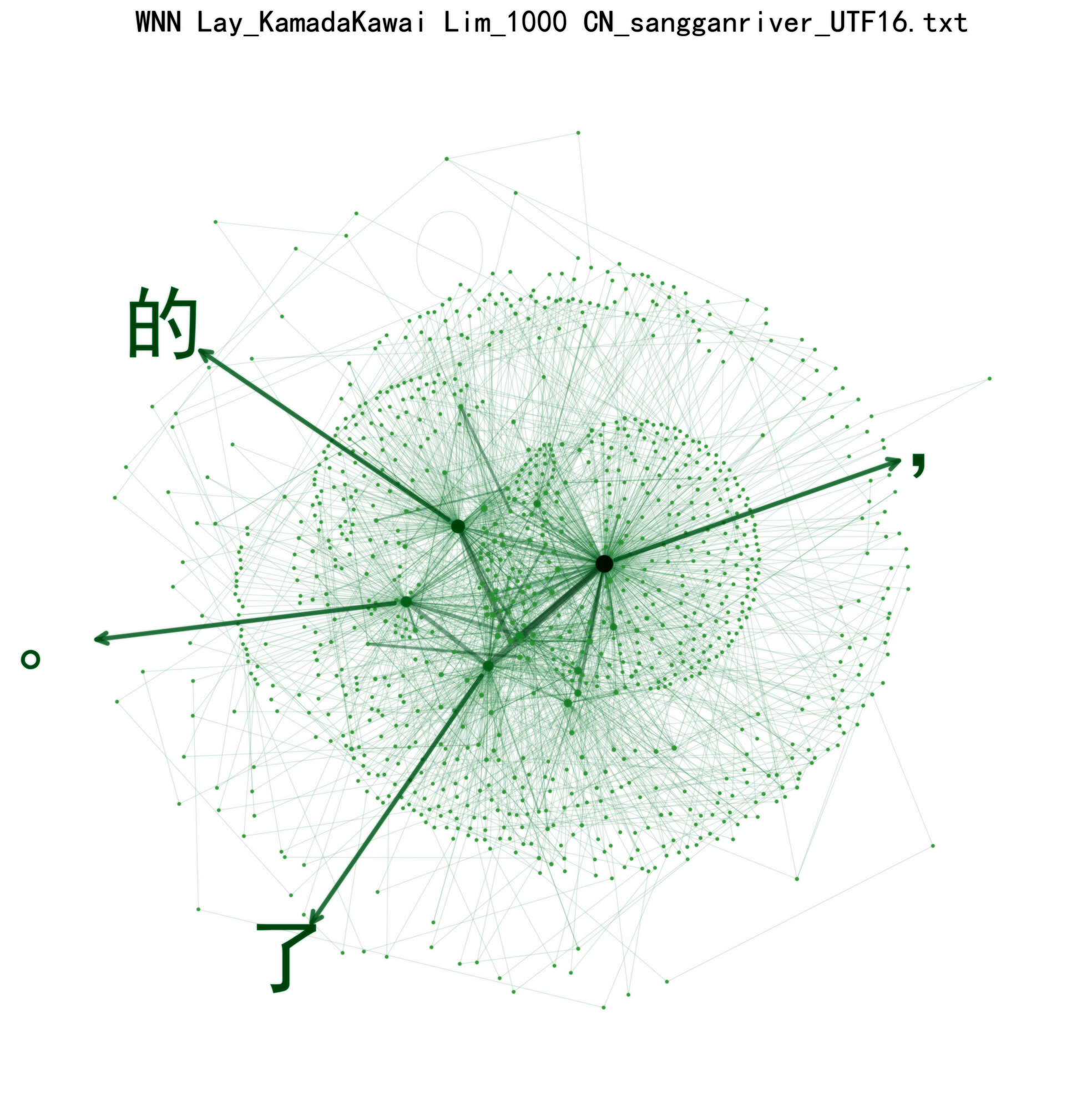}}
\hfill
{\includegraphics[trim={0cm 0 0 1cm}, clip, width = 0.33\linewidth]{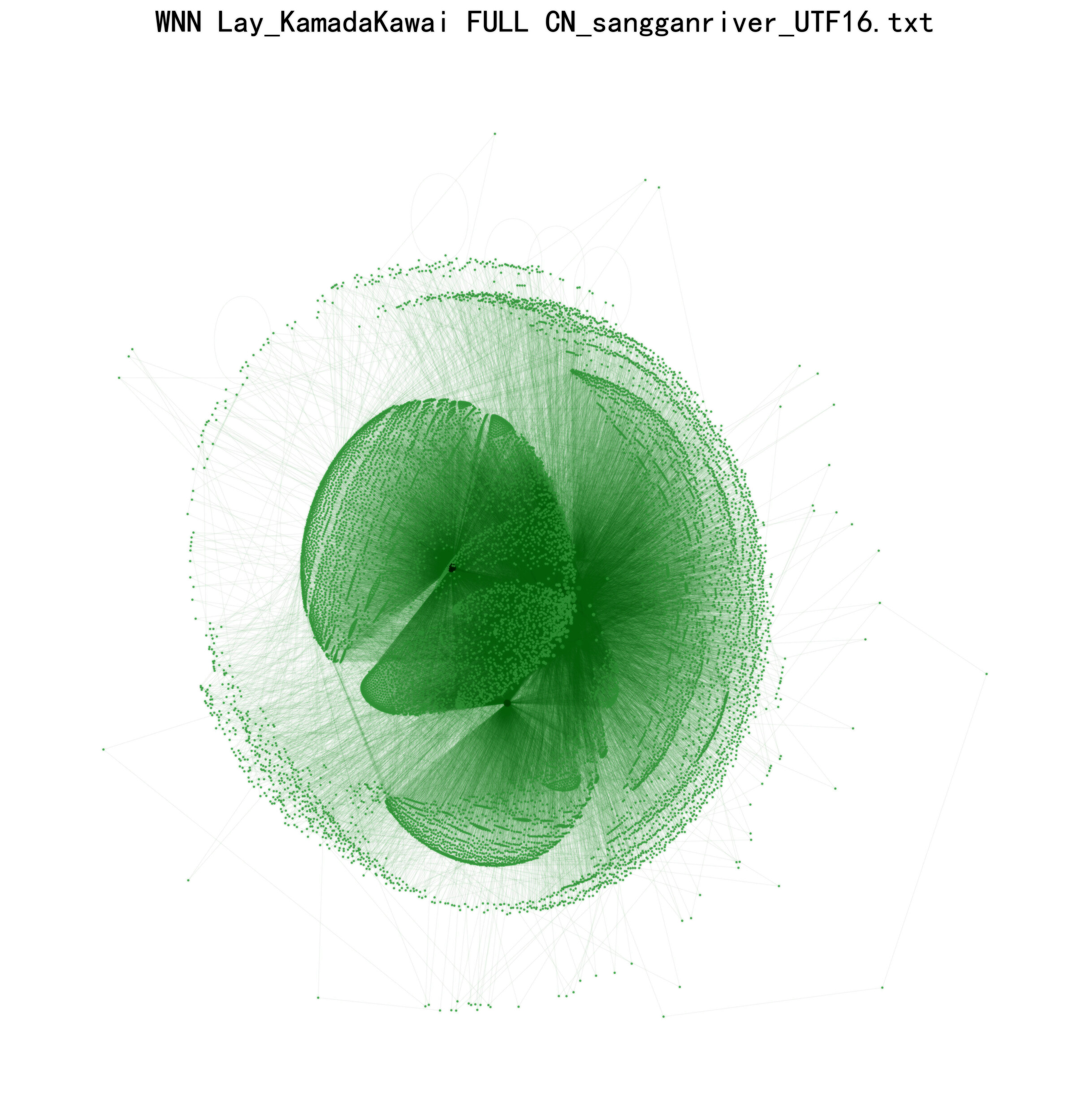}}
\hfill
{\includegraphics[trim={0cm 0 0 0.7cm}, clip, width = 0.32\linewidth]{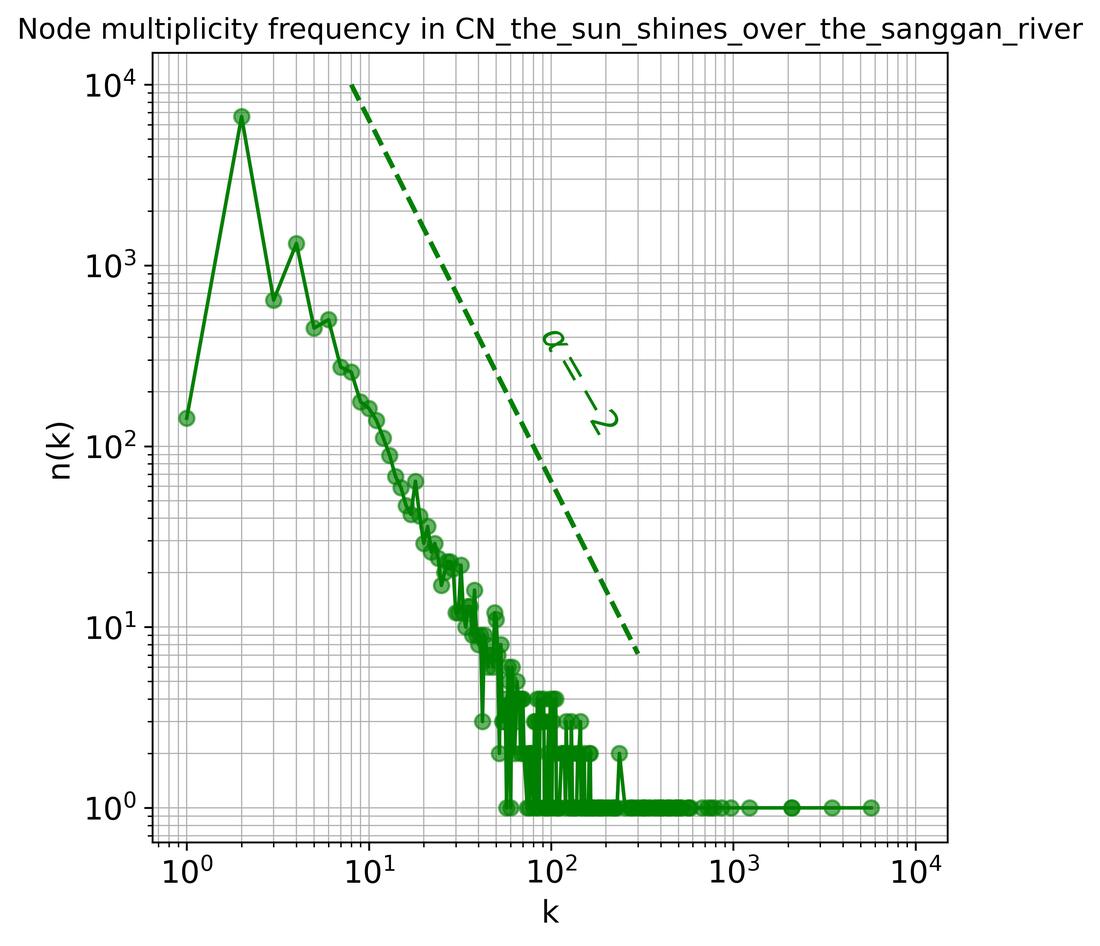}} \\

\textbf{The Sun Shines over the Sanggan River - English} \\
{\includegraphics[trim={0cm 0 0 3.7cm}, clip, width = 0.33\linewidth]{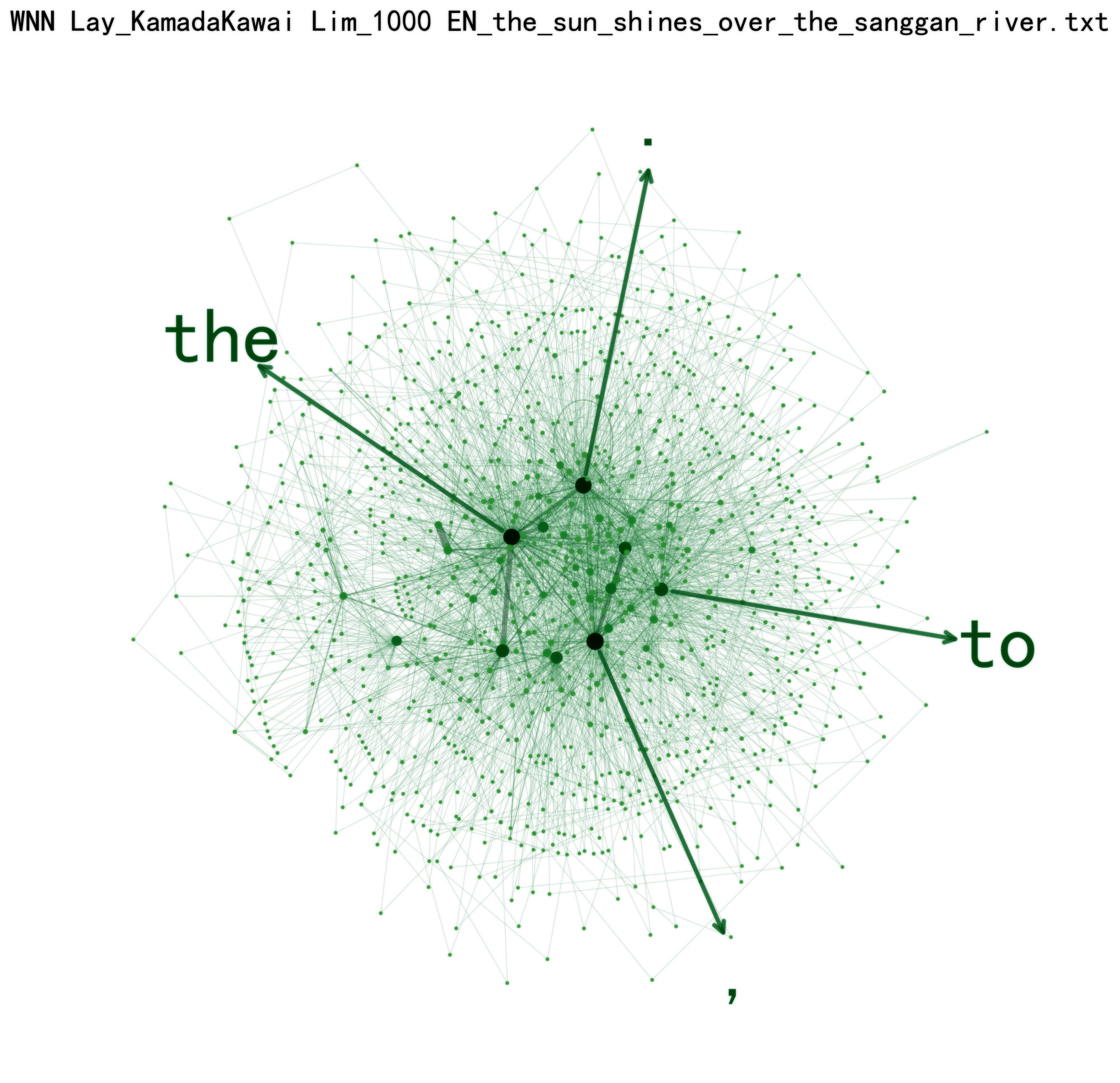}}
\hfill
{\includegraphics[trim={0cm 0 0 0.95cm}, clip, width = 0.33\linewidth]{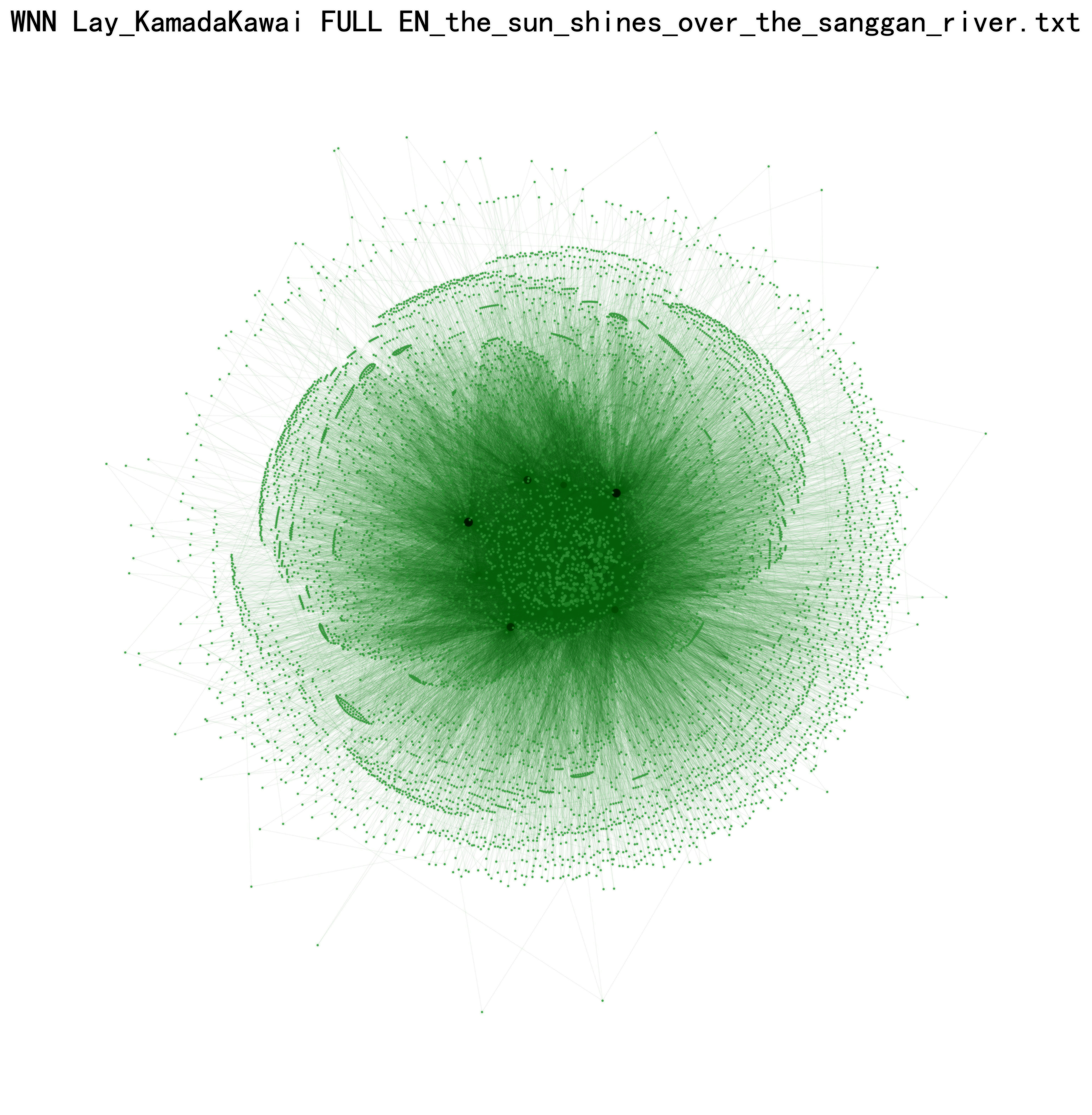}}
\hfill
{\includegraphics[trim={0cm 0 0 0.7cm}, clip, width = 0.32\linewidth]{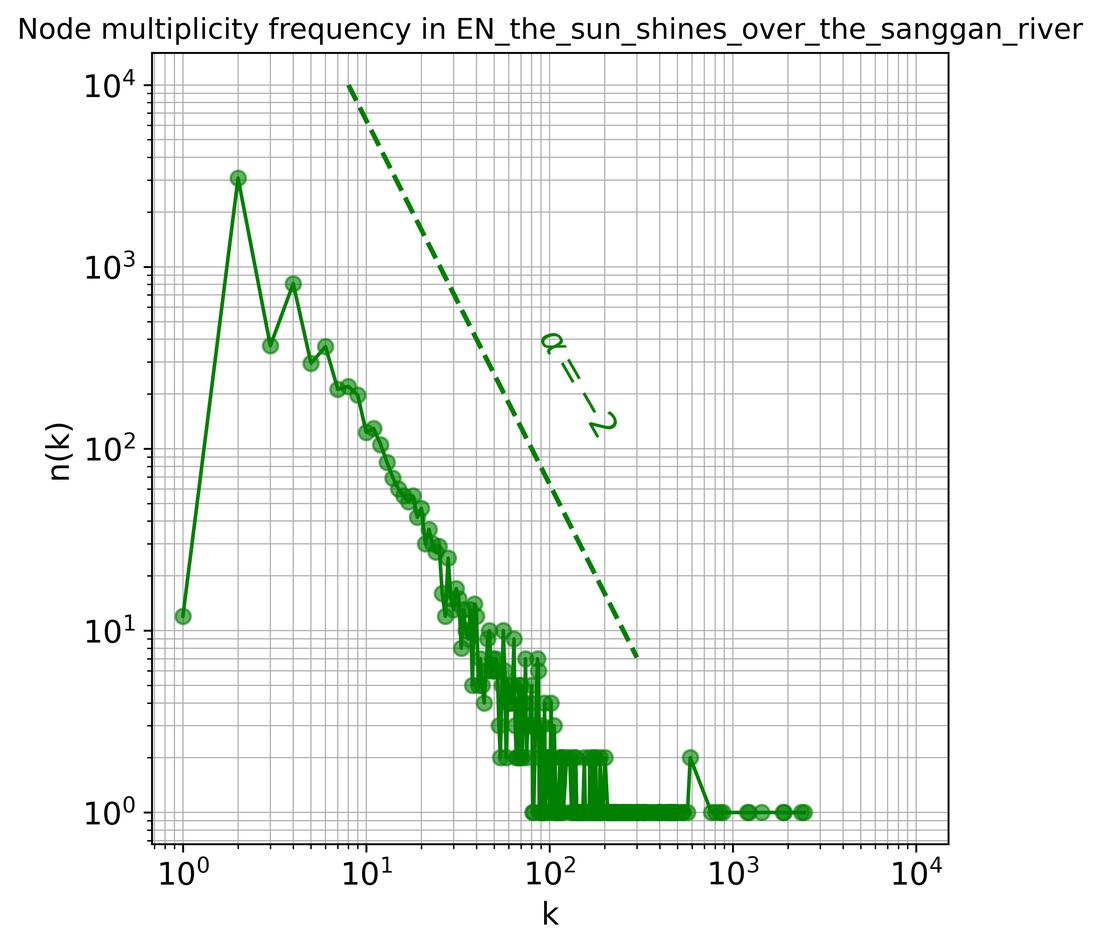}}

\parbox[c]{0.33\linewidth}{\centering First 1000 nodes} \hfill
\parbox[c]{0.33\linewidth}{\centering Entire book} \hfill 
\parbox[b]{0.3\linewidth}{\centering \# $k$-degree nodes}

\vspace{5pt}\hrule\vspace{5pt}
\caption{Word and punctuation-mark adjacency networks for \textit{The Sun Shines over the Sanggan River} in Chinese and English. For each text, the left column corresponds to networks created from the first 1,000 unique words and punctuation marks, the middle column corresponds to networks created from the entire books, and the right column represents the node-degree distributions calculated for the entire books.}
\label{fig::sanggannet}
\vspace{-0.4cm}
\end{figure*}

Figs.~\ref{fig::drunkardnet}--\ref{fig::sanggannet} present the structure of the word-adjacency networks corresponding to selected books originally written in Chinese and their English translations, at two stages of network construction: when the network consists of the first 1000 nodes (roughly corresponding to one chapter of a typical novel), and when the network is complete (i.e., based on the full text). It is clearly visible that each of these networks exhibits a strongly hierarchical structure, with central nodes representing the most frequently used words and punctuation marks, alongside numerous low-degree nodes -- including those of degree 2, which correspond to elements that appear only once in the text. As expected, among the central nodes are those which are associated with the most common punctuation marks in both languages like full stop and comma. Note that the number of nodes in the complete networks representing the studied texts is significant ($10^4 < N_{\rm tot} < 10^6$) and this fact leads to dense network images, in which it is hard to distinguish individual nodes (Figs.~\ref{fig::drunkardnet}--\ref{fig::sanggannet}, middle panels). It is worth noting that these networks are scale-free, and their node degree distributions follow a power law:
\begin{equation}
P(k) \sim k^{-\alpha}
\end{equation}
with an exponent $\alpha \approx 2$ for both the Chinese texts and their translations to English -- see Figs.~\ref{fig::drunkardnet}--\ref{fig::sanggannet} (right panels). This sets these networks apart from those generated by the linear preferential attachment algorithm, which exhibit a power-law exponent of $\alpha = 3$~\cite{BarabasiAL-1999a,AlbertR-2002a,BoccalettiS-2006a}. Also note that in each case one observes a number of outliers, whose degree exceeds that expected for the obtained power laws. On the other side of the distribution, there is a number of nodes with $k=1$. Each of them represent a token that is consistently flanked by another token on both sides throughout the text. To be specific, the constructions like ``to pretend to'', ``to whistle to'', ``as steady as'', and ``and tillie and'' are examples of \textit{hapax logomena} that are connected to only one token in \textit{Alice's Adventures in Wonderland}. The number of such token triples is relatively small and the respective nodes do not comply with the overall power law. A comparison of the degrees of the largest hubs in networks constructed from original Chinese texts and their English translations reveals that the maximum node degree tends to be slightly lower in the translated English versions. This difference can be attributed to the structural characteristics of the Chinese language, in which function words (like \begin{CJK}{UTF8}{gbsn}的\end{CJK} `de', \begin{CJK}{UTF8}{gbsn}被\end{CJK} `bèi', \begin{CJK}{UTF8}{gbsn}了\end{CJK} `le', \begin{CJK}{UTF8}{gbsn}把\end{CJK} `bǎ', etc.) and other words (like \begin{CJK}{UTF8}{gbsn}一\end{CJK}`yī', \begin{CJK}{UTF8}{gbsn}是\end{CJK} `shì', \begin{CJK}{UTF8}{gbsn}在\end{CJK} `zài', etc.) occur relatively more frequently and exert a stronger influence on the topology of the word-adjacency networks than is the case in English. It should also be noted that, due to the lack of lemmatization of English words in this analysis, verbs and nouns are often split across multiple nodes corresponding to their various inflectional forms - a phenomenon that does not occur in contemporary Chinese, where inflection is absent and its role is expressed by word order and particles~\cite{ZhangL-2023a}.

\begin{figure*}
\centering
\begin{minipage}[t]{0.48\linewidth}
\centering
(a)\includegraphics[width=\linewidth]{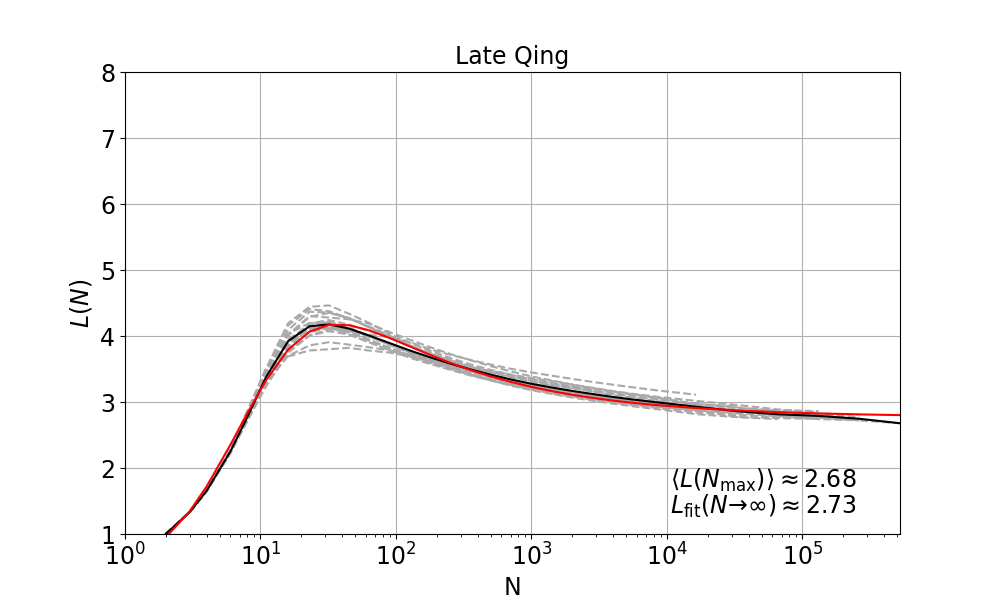}
\end{minipage}
\hfill
\begin{minipage}[t]{0.48\linewidth}
\centering
(b)\includegraphics[width=\linewidth]{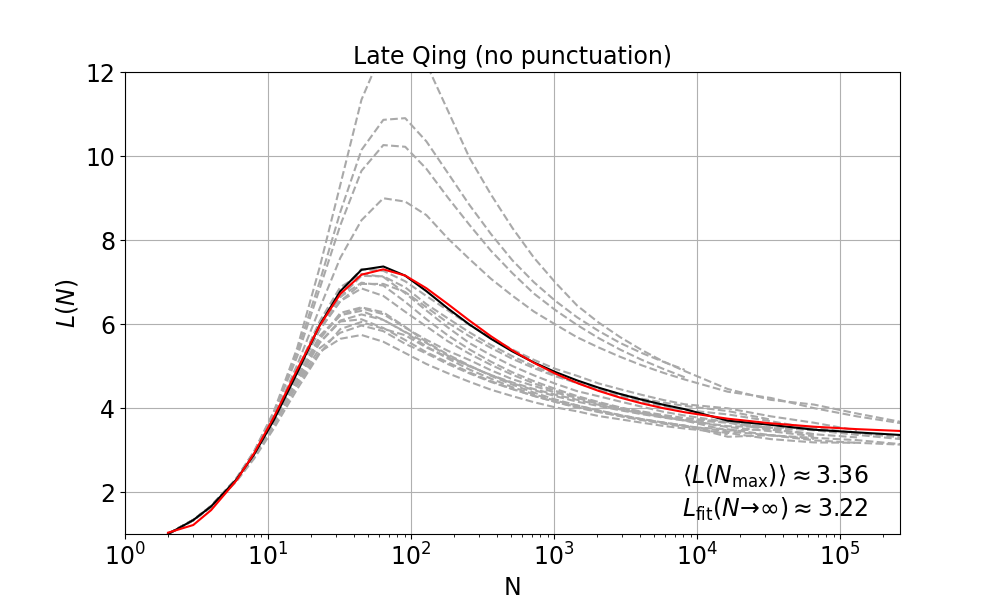}
\end{minipage}

\begin{minipage}[t]{0.48\linewidth}
\centering
(c)\includegraphics[width=\linewidth]{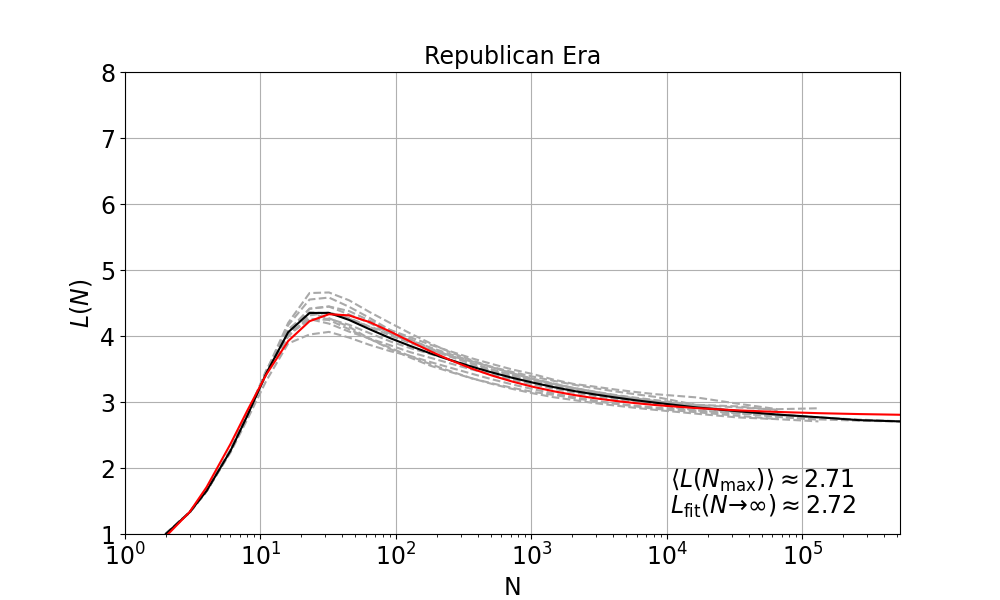}
\end{minipage}
\hfill
\begin{minipage}[t]{0.48\linewidth}
\centering
(d)\includegraphics[width=\linewidth]{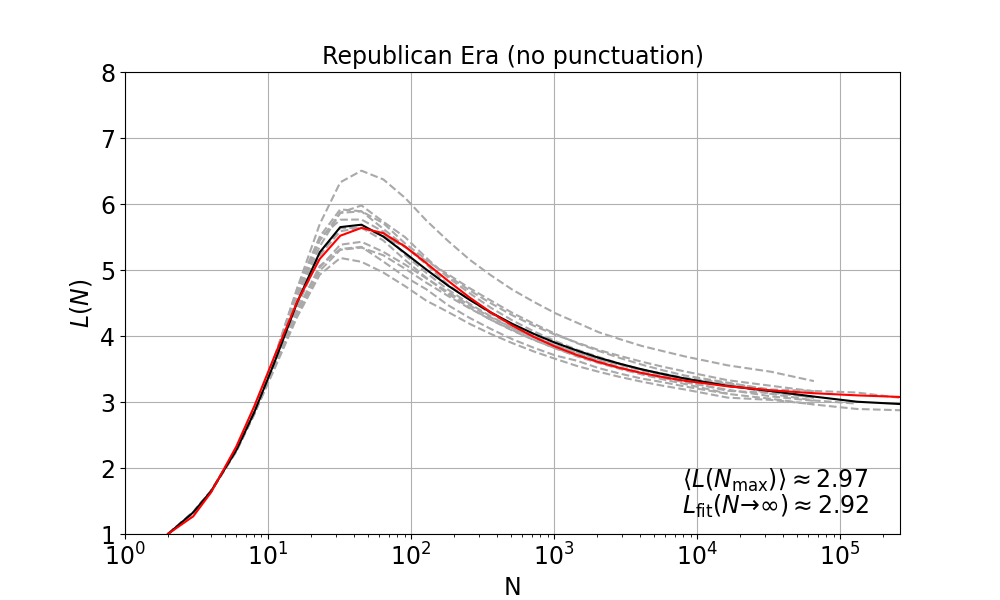}
\end{minipage}

\begin{minipage}[t]{0.48\linewidth}
\centering
(e)\includegraphics[width=\linewidth]{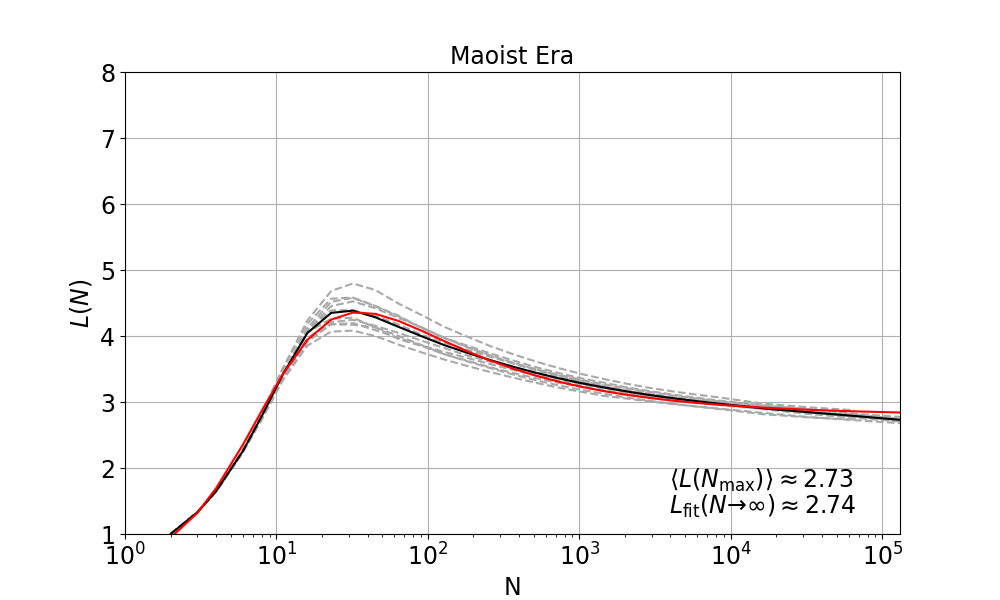}
\end{minipage}
\hfill
\begin{minipage}[t]{0.48\linewidth}
\centering
(f)\includegraphics[width=\linewidth]{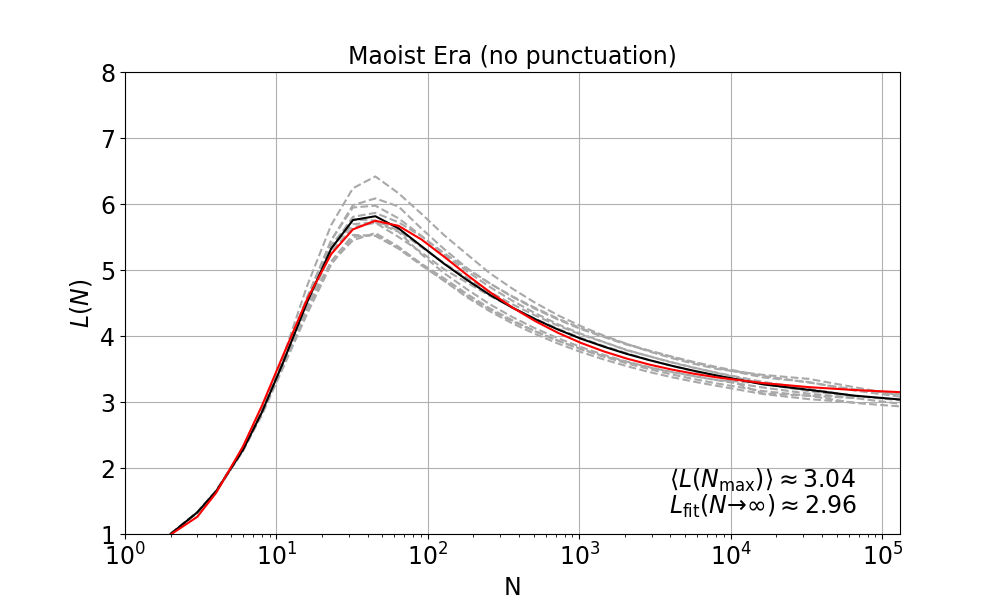}
\end{minipage}

\begin{minipage}[t]{0.48\linewidth}
\centering
(g)\includegraphics[width=\linewidth]{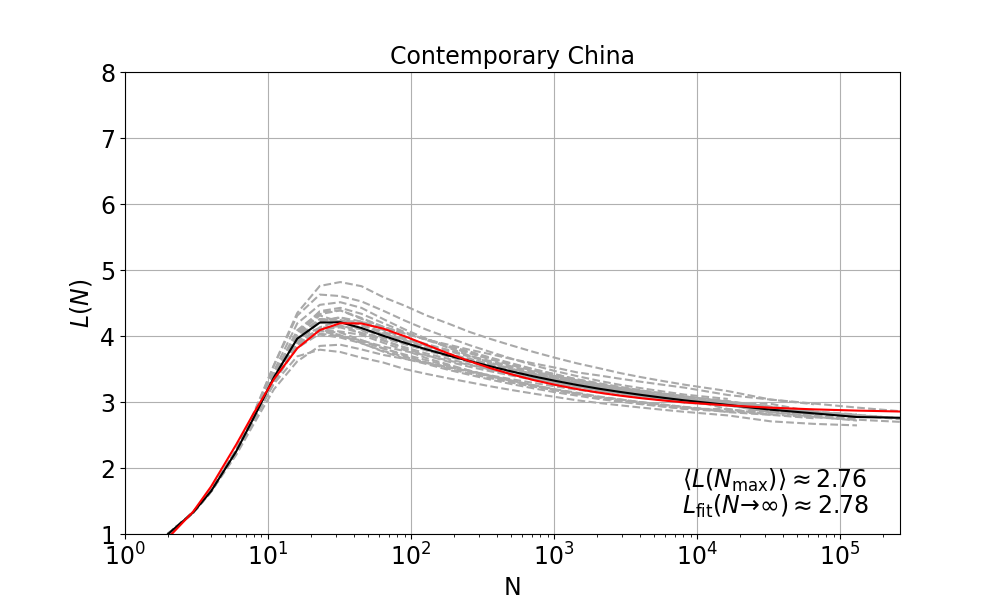}
\end{minipage}
\hfill
\begin{minipage}[t]{0.48\linewidth}
\centering
(h)\includegraphics[width=\linewidth]{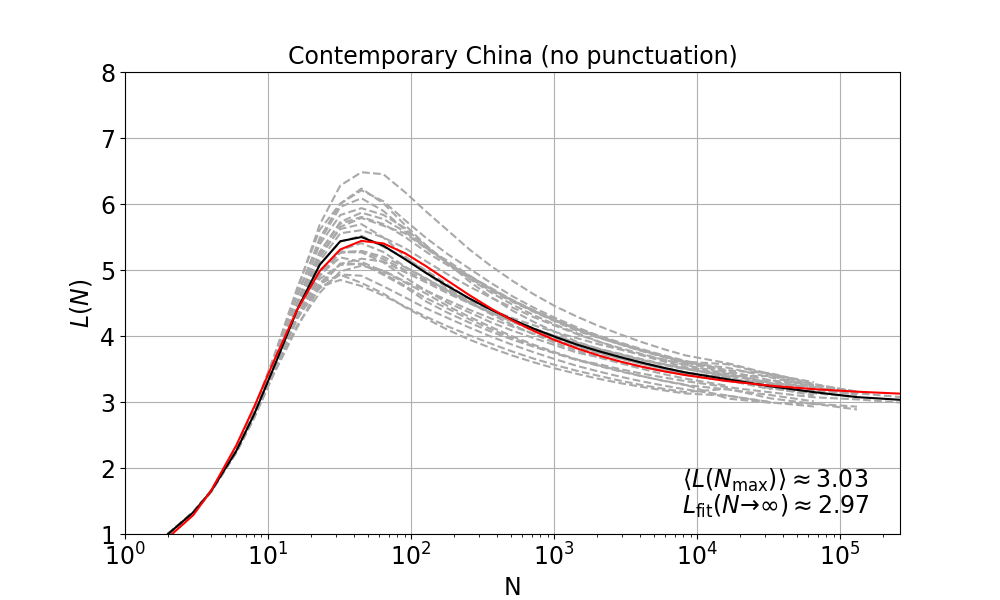}
\end{minipage}
\caption{Average shortest path length $L(N)$ as a function of network size $N$ for a set of individual texts (grey dashed lines) together with mean ASPL $\langle L(N) \rangle$ averaged over all the texts (solid black) and the model fitted to it (solid red). Three sets are shown: 18 novels from the Late Qing era and early republican era till 1911 (top), 11 novels from the republican era 1912-1948 (upper middle), 10 novels from the Maoist era 1949-1978 (lower middle), and 23 contemporary Chinese novels since 1979 (bottom). The texts with punctuation marks (left) and without punctuation marks (right) are shown separately. Each empirical function has been averaged over networks obtained by shifting the starting point of the text by a fixed number of words. Note an extended scale of the vertical axis in (b).}
\label{fig::chinese.corpora.1}
\vspace{-0.4cm}
\end{figure*}

Fig.~\ref{fig::chinese.corpora.1} shows the behavior of the average shortest path length as a function of network size for each Chinese book in the dataset, presented in two variants: networks constructed from words and punctuation marks treated equally with words (left column) and networks composed solely of words, with punctuation ignored -- two words can be neighbors even if they are separated by a punctuation mark (right column). To avoid the situation, in which the behavior of ASPL at the initial stage of network construction ((i.e., for small values of $N$) is overly influenced by the specific selection of words and punctuation marks at the beginning of the text~\cite{KuligA-2015a}, the $L(N)$ function was averaged over multiple realizations with varying starting points. Each text was concatenated into a closed cycle, and the token or word, from which network construction began, was shifted forward by an increment of $\Delta \tau$ tokens/words in each realization. Due to the substantial decrease in performance of the ASPL computation algorithm for large $N$, the number of such realizations was gradually reduced as $N$ increased. In particular, for $N \leqslant 10{,}000$, the step size was $\Delta \tau = 100$, for $10{,}000 < N \leqslant 100{,}000$, the shift increased to $\Delta \tau = 1000$, and so on. Therefore, all the results presented in this work were calculated with such averaging.

An immediately noticeable feature of these plots is the increase in ASPL, which is the most significant in the central part of the plots ($10 < N < 1000$), if punctuation has been ignored. This is effectively equivalent to removing a number of nodes from the networks on the left, including some hubs, which naturally leads to a partial decentralization of the network and to an increase in ASPL. Typically, the increase ranges from about 4 to about 5 or 6, but exceptions do occur. In the most extreme case, the maximum of $L(N)$ increases from 3.7 to 12.8 for the book \textit{Eternal Happiness}, published during the LQ period. Overall, books from that era exhibit a stronger impact of punctuation on the network topology compared to those from other periods under study -- Fig.~\ref{fig::chinese.corpora.1}(b). During this era, the highest degree of variation in maximum values across texts without punctuation is observed. The smallest variation occurs for ME, at least in the considered set of texts. In contrast, the behavior of $L(N)$ for the networks that include punctuation is considerably more homogeneous as Fig.~\ref{fig::chinese.corpora.1}(a)(c)(e)(g) documents. This is even better visible if one calculates the average $L(N)$ from the texts within each era and look at its maximum: $\langle L \rangle_{\rm max}$ remains almost invariant under time translation if punctuation is considered. This invariance disintegrates after punctuation has been neglected, though, and the maximum of the average ASPL changes between the eras: $5.5 < \langle L \rangle_{\rm max} < 7.5$.

\begin{figure*}
\centering
\begin{minipage}[t]{0.48\linewidth}
\centering
(a)\includegraphics[width=\linewidth]{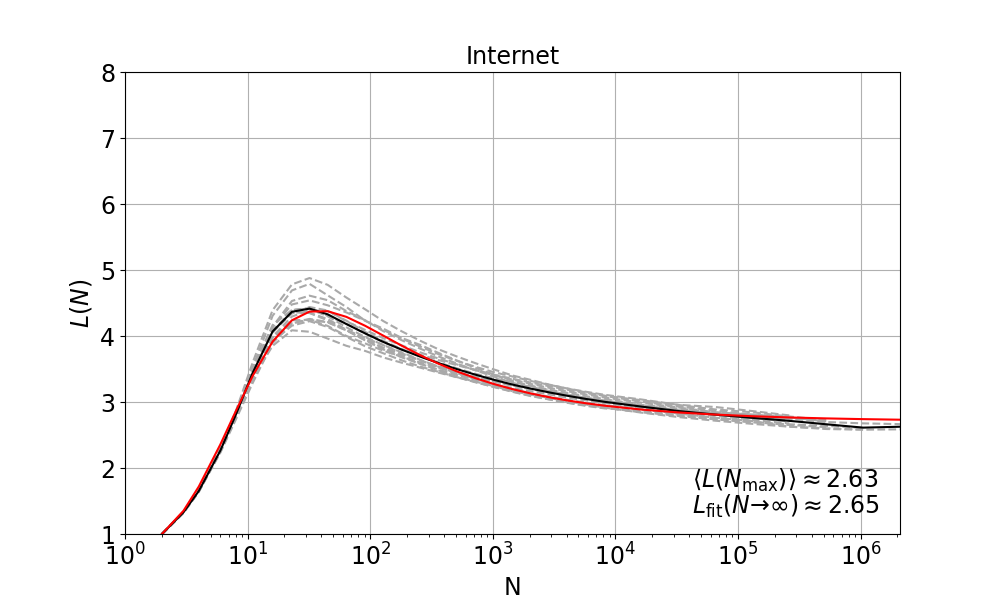}
\end{minipage}
\hfill
\begin{minipage}[t]{0.48\linewidth}
\centering
(b)\includegraphics[width=\linewidth]{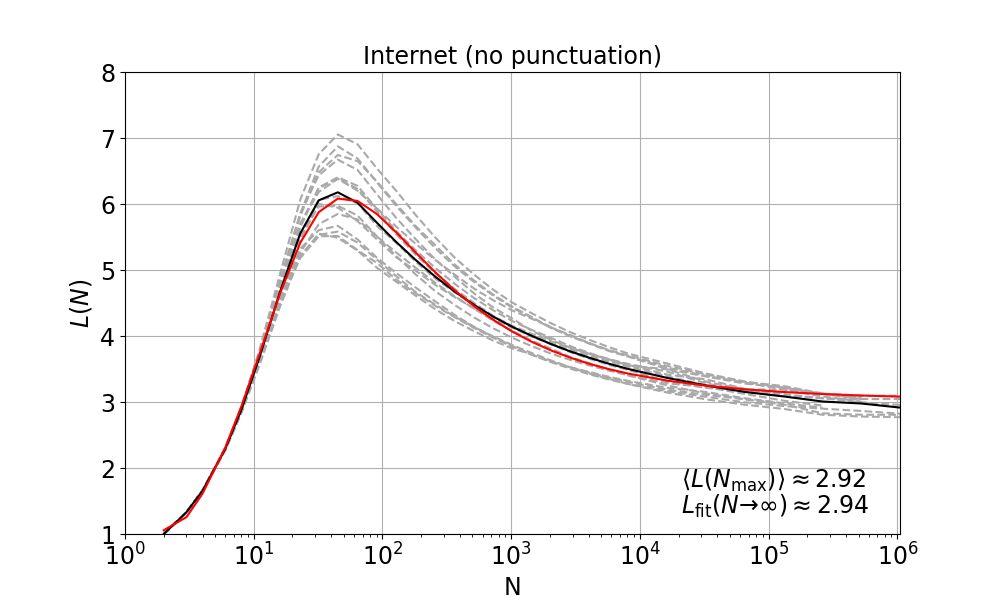}
\end{minipage}

\begin{minipage}[t]{0.48\linewidth}
\centering
(c)\includegraphics[width=\linewidth]{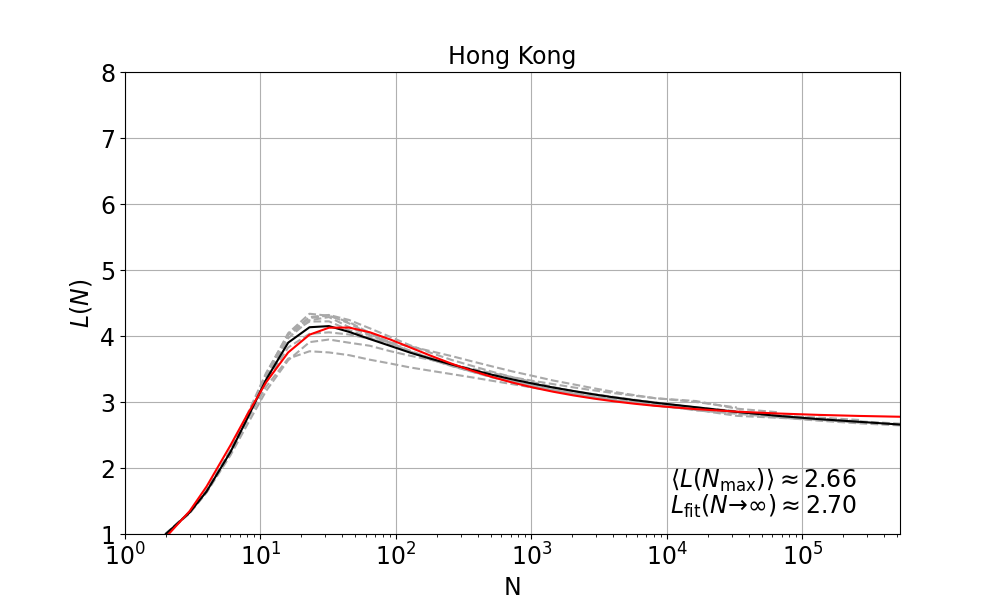}
\end{minipage}
\hfill
\begin{minipage}[t]{0.48\linewidth}
\centering
(d)\includegraphics[width=\linewidth]{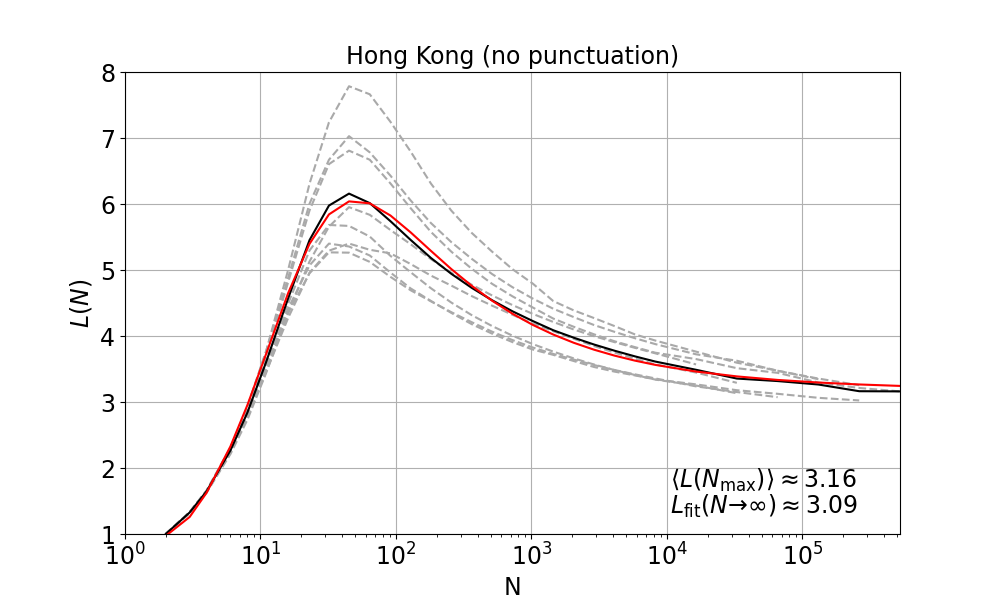}
\end{minipage}

\begin{minipage}[t]{0.48\linewidth}
\centering
(e)\includegraphics[width=\linewidth]{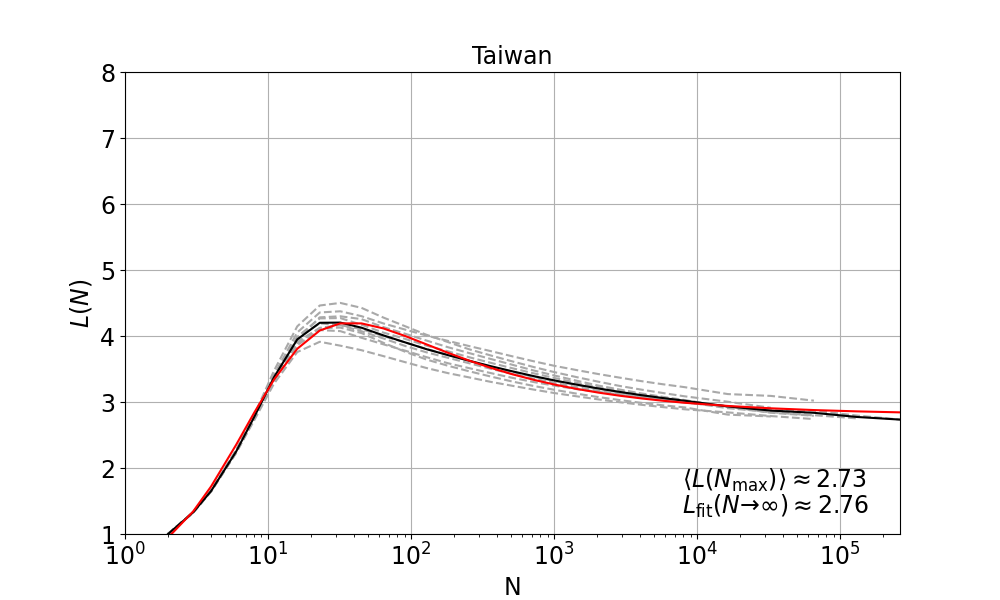}
\end{minipage}
\hfill
\begin{minipage}[t]{0.48\linewidth}
\centering
(f)\includegraphics[width=\linewidth]{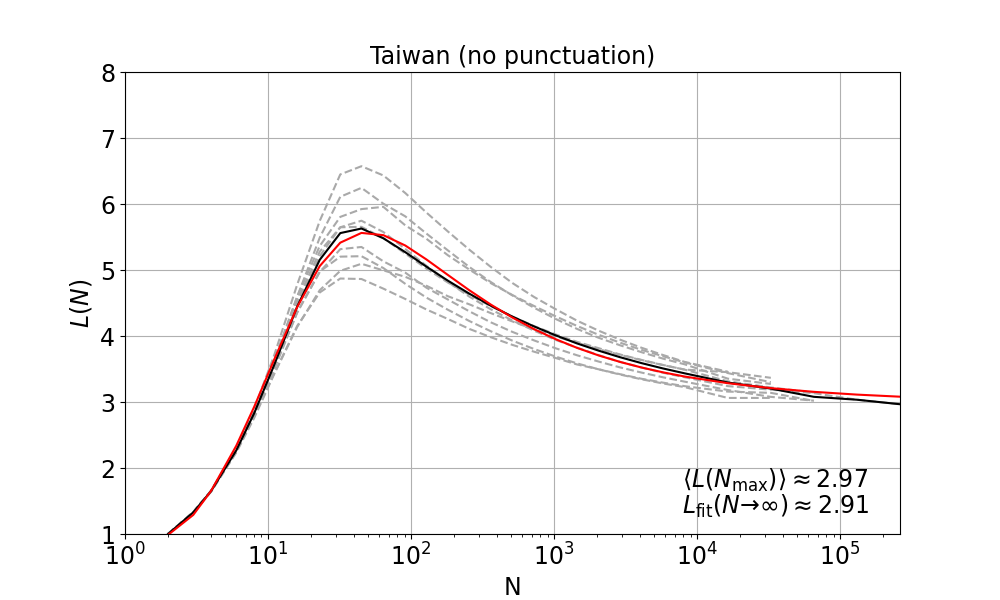}
\end{minipage}
\caption{Average shortest path length $L(N)$ as a function of network size $N$ for a set of individual texts (grey dashed lines) together with the ASPL averaged over all the texts (solid black) and the hybrid model fitted to the average ASPL (solid red). Three sets are shown: 15 Internet novels (top), 8 novels written by Hong Kong writers (middle), and 9 novels written by Taiwanese writers (bottom). The texts with punctuation marks (left) and without punctuation marks (right) are shown separately. Each empirical function has been averaged over networks obtained by shifting the starting point of the text by a fixed number of words.}
\label{fig::chinese.corpora.2}
\vspace{-0.4cm}
\end{figure*}

\begin{figure*}
\centering
\begin{minipage}[t]{0.48\linewidth}
\centering
(a)\includegraphics[width=\linewidth]{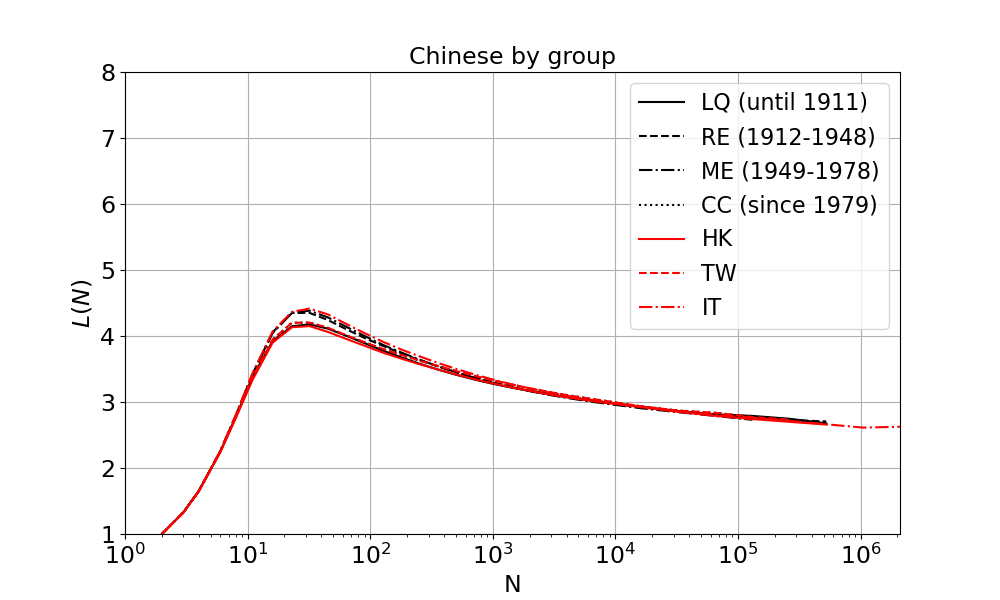}
\end{minipage}
\hfill
\begin{minipage}[t]{0.48\linewidth}
\centering
(b)\includegraphics[width=\linewidth]{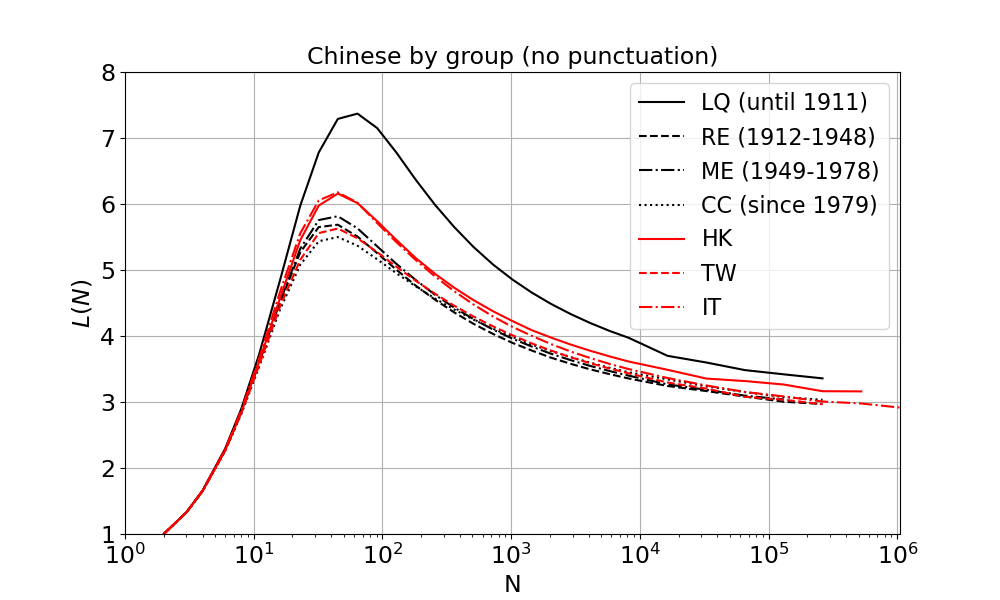}
\end{minipage}
\caption{Mean average shortest path length $\langle L(N) \rangle$ as a function of network size $N$ for different subsets of Chinese texts: Late Qing Era (LQ, 18 novels), Republican Era (RE, 11 novels), Maoist Era (ME, 10 novels), Contemporary China (CC, 23 novels), Hong Kong (HK, 8 novels), Taiwan (TW, 9 novels), and Internet (IT, 15 novels). The texts with punctuation marks (left) and without punctuation marks (right) are shown separately.}
\label{fig::chinese.corpora.allgroups}
\vspace{-0.4cm}
\end{figure*}

In addition to the collections comprising texts from various literary periods in mainland China, the analysis also included texts written by authors from Taiwan, Hong Kong, and by authors publishing online. In the latter case, the texts are significantly longer - each containing several hundred thousand tokens. Fig.~\ref{fig::chinese.corpora.2} presents the results grouped analogously to Fig.~\ref{fig::chinese.corpora.1}. Here, the differences between individual texts in terms of the maximum value of the average ASPL for the networks without punctuation are larger than those observed in mainland Chinese literature outside the LQ era, but still smaller than those found in that specific era. In both the case with punctuation included and the case where it is omitted, the maximum value of $\langle L \rangle_{\rm max}$ does not significantly differ from what was observed above. The comparison of ASPL behavior between the different text groups, presented above, becomes even easier when looking at Fig.~\ref{fig::chinese.corpora.allgroups}, where the average ASPL values for each group are overlaid for the case of token networks (Fig.~\ref{fig::chinese.corpora.allgroups}(a)) and word networks (Fig.~\ref{fig::chinese.corpora.allgroups}(b)).

We now focus on the behavior of $L(N)$ in the limit of very large networks, specifically examining the asymptotic behavior $\lim_{N \to \infty} L(N)$. For large values of $N$, word co-occurrence networks are characterized by the average ASPL that decreases with increasing $N$. This behavior can be approximated, within a certain range, by the function given in Eq.(\ref{eq::aspl.large.network}), with the asymptotic value defined in Eq.~(\ref{eq::aspl.asymptotic}). The fit to $\langle L(N) \rangle$ is shown in each panel of Figs.~\ref{fig::chinese.corpora.1} and~\ref{fig::chinese.corpora.2}. It can be seen clearly that the fit remains good up to around $N \approx 100{,}000$, beyond which the empirical average becomes noticeably smaller than the model predictions. However, this difference is not large and is bounded by $\Delta L = L_{\rm fit}(N_{\rm tot}) - \langle L(N_{\rm tot}) \rangle \lessapprox 0.05$. The reason for the observed inaccuracy of the model can be explained by considering that empirical word/token adjacency networks are not the Erdős-Rényi-type random graphs and exhibit a centralized structure, although significantly less so than, for example, Barab\'asi-Albert networks. However, due to the fact that the selection of nodes receiving new edges during each step of the network growth only gradually approaches linear preferential attachment for large $N$, the resulting deviation reflects a transition from the earlier, more uniform (``democratic'') growth behavior toward a more centralized one. The empirical value of $\langle L(N_{\rm tot}) \rangle$ and the asymptotic value of $L_{\rm fit}(N)$ are shown in each panel of both Figures.

In the case of the token adjacency networks, the asymptotic values of the mean ASPL derived from the fitted model for all text subsets except for the Internet novels exhibit a relatively narrow range, from 2.70 to 2.78, which suggests that, for Chinese texts, the network topology as captured by this measure is universal and depends on neither the literary era nor the place of living of the authors. A smaller value of the asymptotic for the Internet texts ($L_{\rm fit}(N \to \infty) = 2.65$) comes from the fact that they include a broader spectrum of available characters (like emojis), which effectively shorten the inter-node paths. A slightly greater variability in $L_{\rm fit}(N)$ is observed for networks based solely on word adjacency, and the deviations between the fitted model and empirical values are somewhat larger. This can indicate that the most frequently occurring punctuation marks, by acting as significant hubs, contribute to the homogenization of the network topology.

\begin{figure*}
\centering
\begin{minipage}[t]{0.48\linewidth}
\centering
(a)\includegraphics[width=\linewidth]{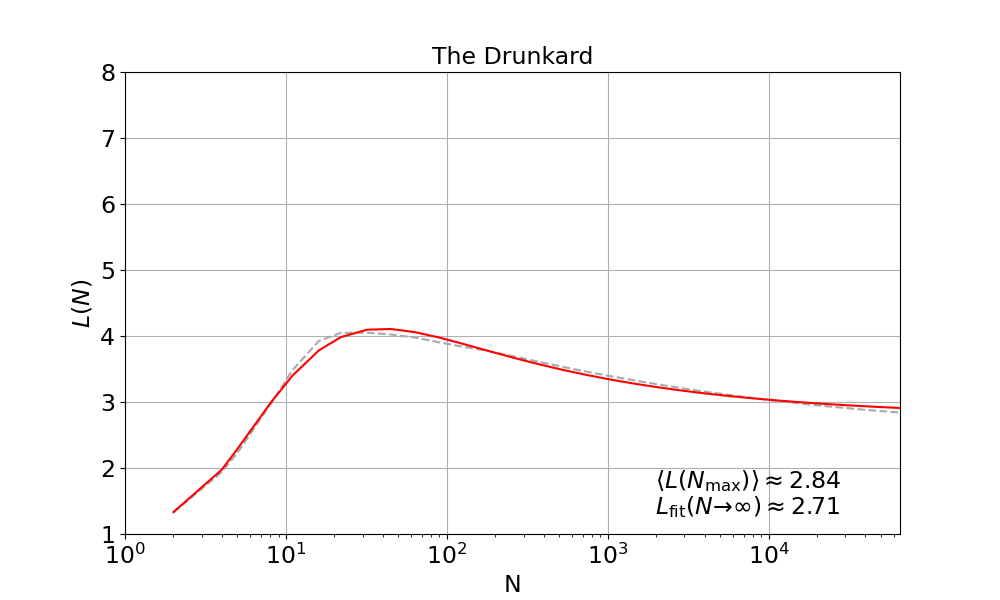}
\end{minipage}
\hfill
\begin{minipage}[t]{0.48\linewidth}
\centering
(b)\includegraphics[width=\linewidth]{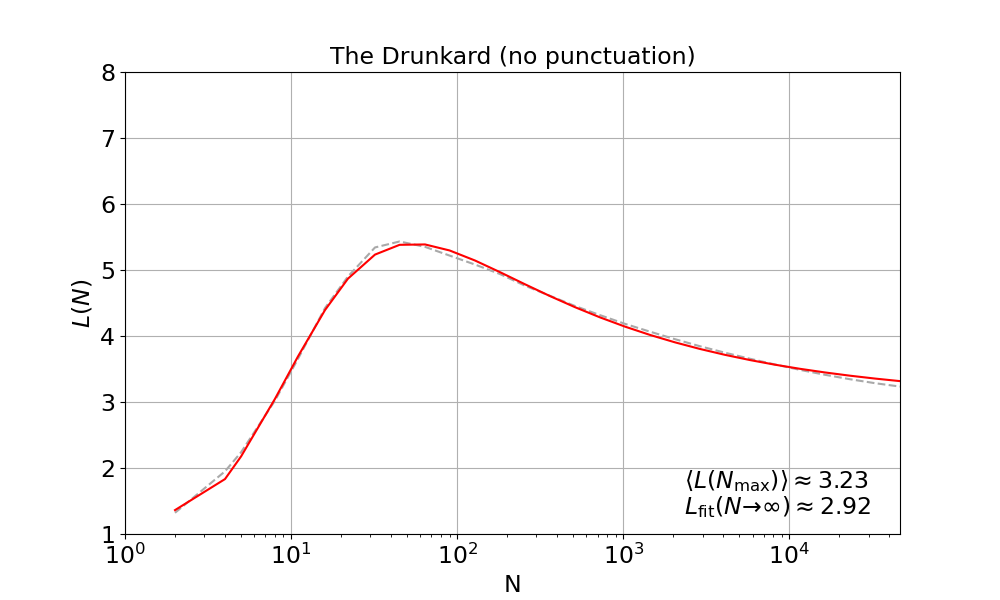}
\end{minipage}

\begin{minipage}[t]{0.48\linewidth}
\centering
(c)\includegraphics[width=\linewidth]{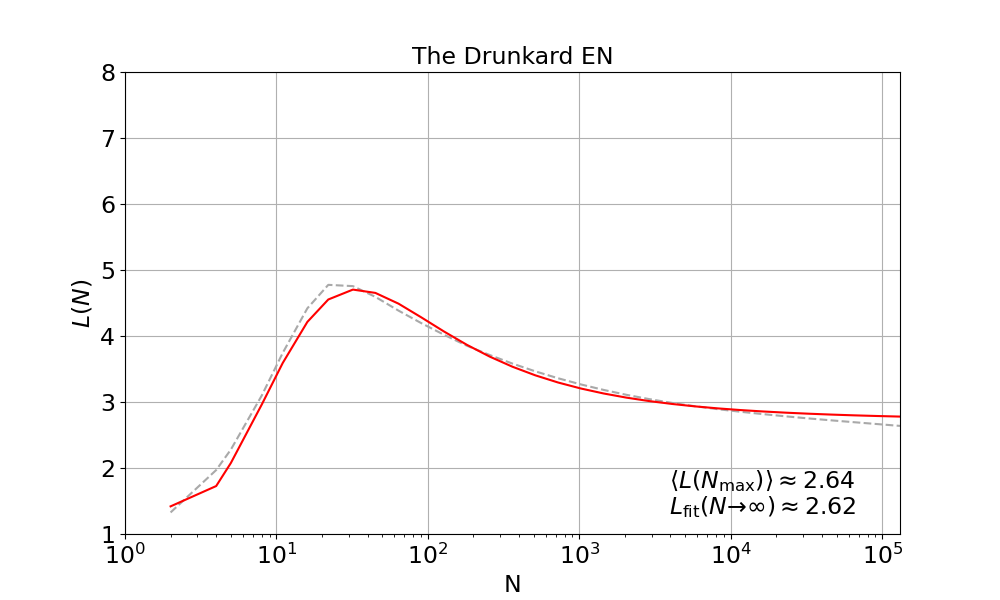}
\end{minipage}
\hfill
\begin{minipage}[t]{0.48\linewidth}
\centering
(d)\includegraphics[width=\linewidth]{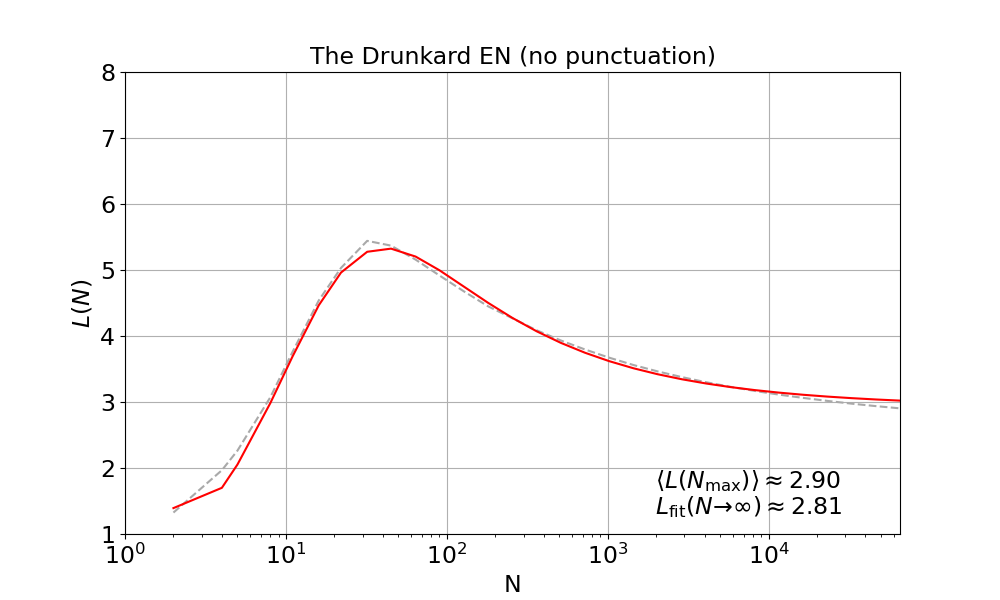}
\end{minipage}
\caption{Average shortest path length $L(N)$ as a function of network size $N$ for \textit{Drunkard} by Liu Yi-Chang (dashed) together with the fitted hybrid model (solid). Chinese original version with punctuation marks (a) and without punctuation marks (b) is shown together with its English translation with (c) and without punctuation marks (d). For each text, numerous growing network realizations were created by shifting the text starting point by a given number of words; the presented $L(N)$ was calculated as an average over all these realizations.}
\label{fig::drunkard}
\vspace{-0.4cm}
\end{figure*}

\begin{figure*}
\centering
\begin{minipage}[t]{0.48\linewidth}
\centering
(a)\includegraphics[width=\linewidth]{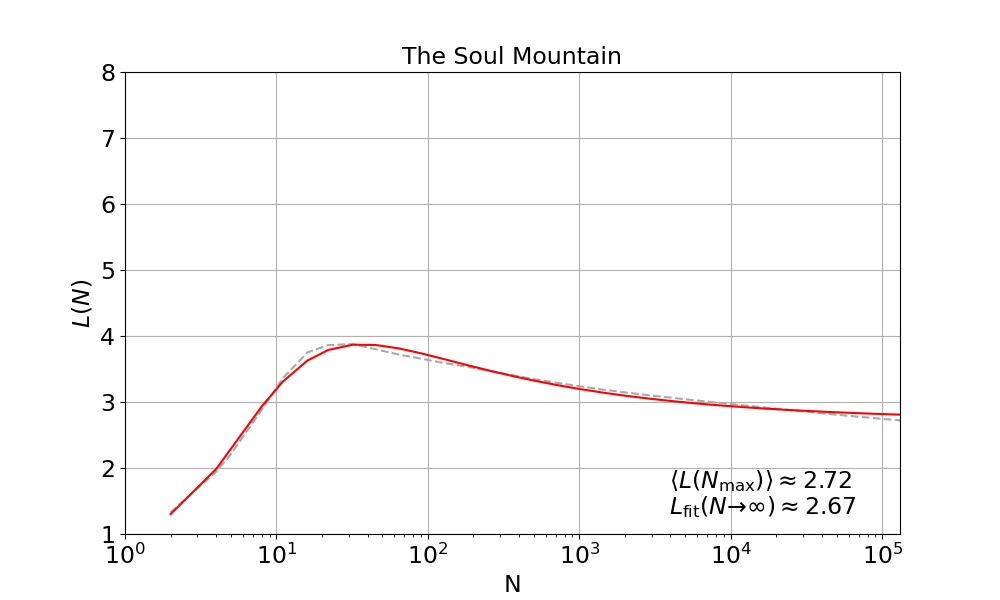}
\end{minipage}
\hfill
\begin{minipage}[t]{0.48\linewidth}
\centering
(b)\includegraphics[width=\linewidth]{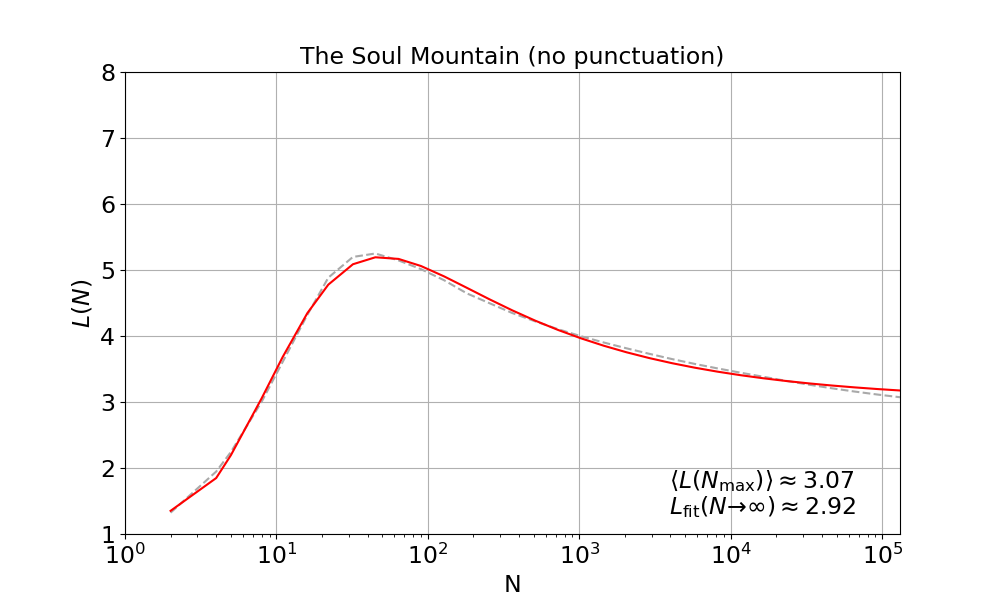}
\end{minipage}

\begin{minipage}[t]{0.48\linewidth}
\centering
(c)\includegraphics[width=\linewidth]{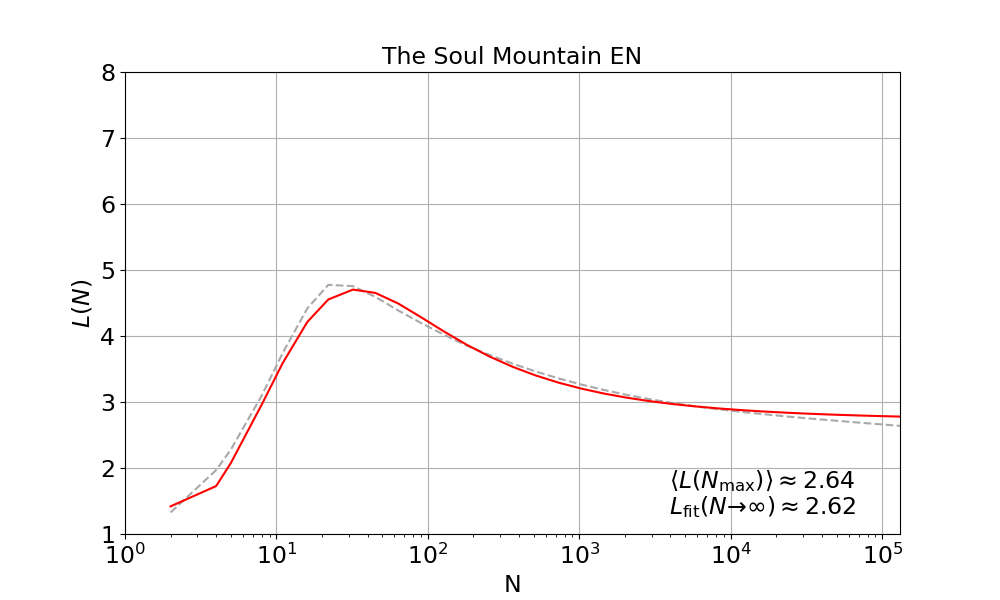}
\end{minipage}
\hfill
\begin{minipage}[t]{0.48\linewidth}
\centering
(d)\includegraphics[width=\linewidth]{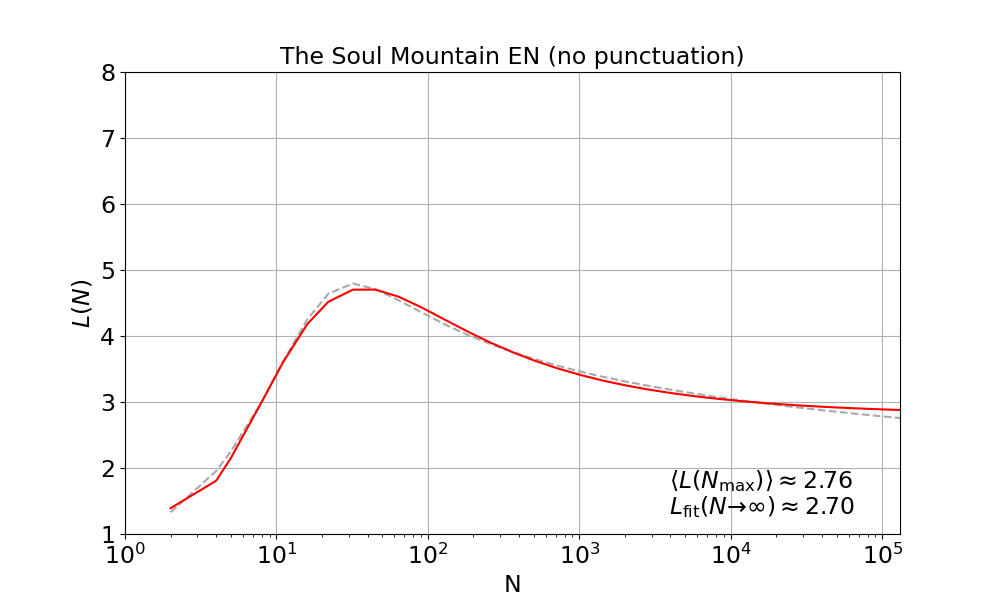}
\end{minipage}
\caption{Average shortest path length $L(N)$ as a function of network size $N$ for \textit{Soul Mountain} by Gao Xingjian (dashed) together with the fitted hybrid model (solid). Chinese original version with punctuation marks (a) and without punctuation marks (b) is shown together with its English translation with (c) and without punctuation marks (d). For each text, numerous growing network realizations were created by shifting the text starting point by a given number of words; the presented $L(N)$ was calculated as an average over all these realizations.}
\label{fig::soul_mountain}
\vspace{-0.4cm}
\end{figure*}

\begin{figure*}
\centering
\begin{minipage}[t]{0.48\linewidth}
\centering
(a)\includegraphics[width=\linewidth]{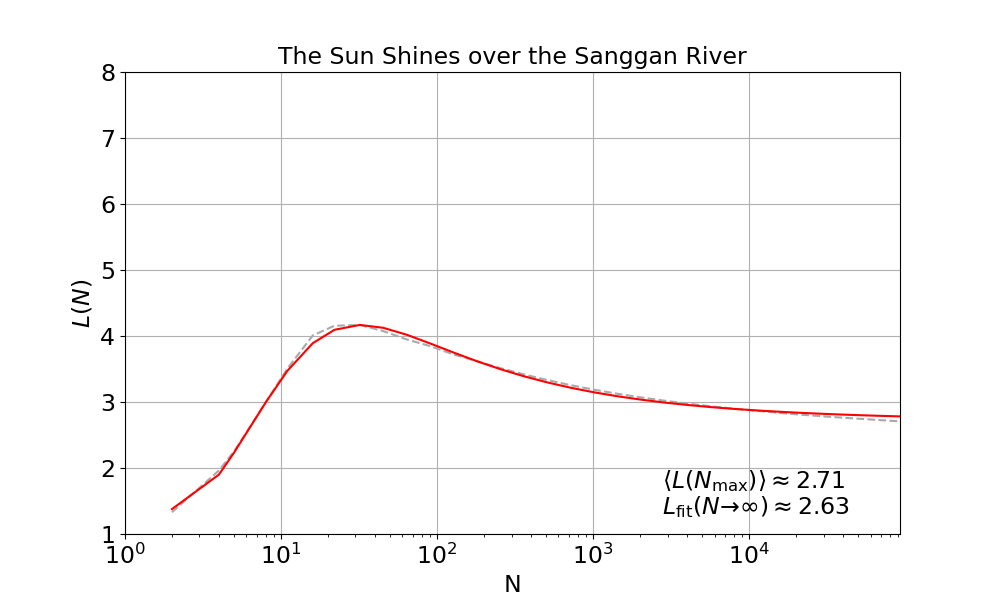}
\end{minipage}
\hfill
\begin{minipage}[t]{0.48\linewidth}
\centering
(b)\includegraphics[width=\linewidth]{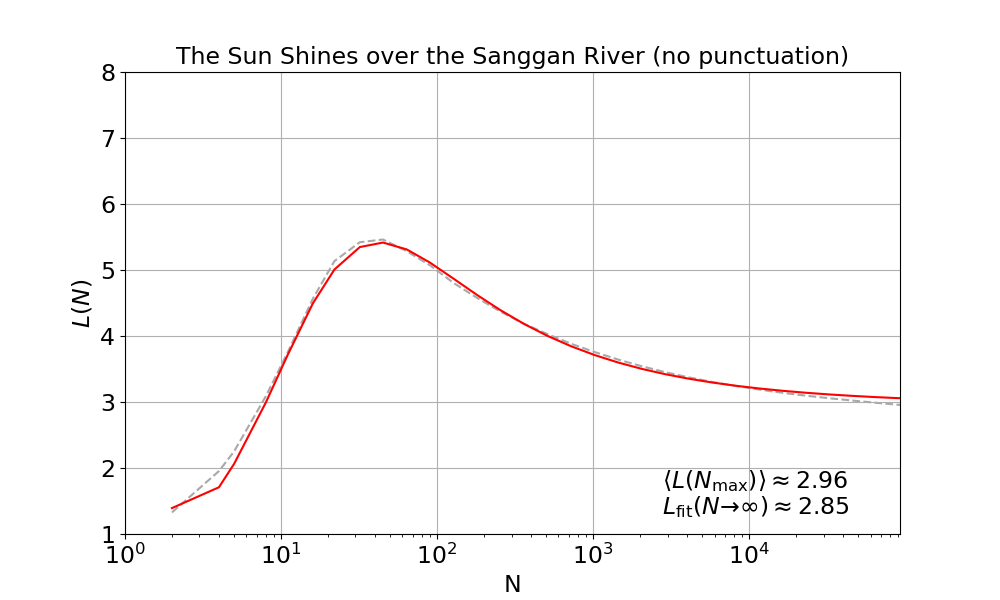}
\end{minipage}

\begin{minipage}[t]{0.48\linewidth}
\centering
(c)\includegraphics[width=\linewidth]{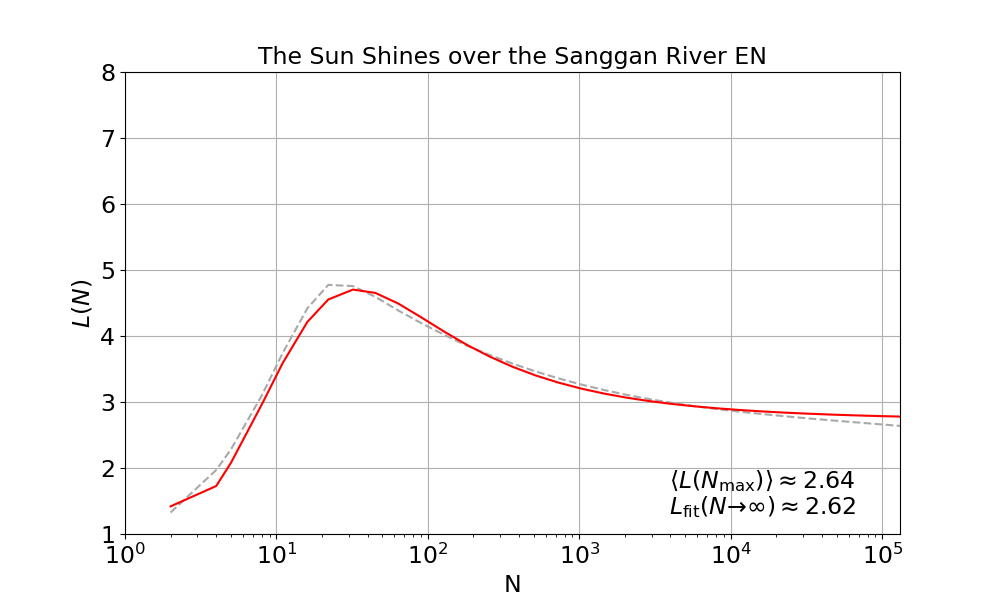}
\end{minipage}
\hfill
\begin{minipage}[t]{0.48\linewidth}
\centering
(d)\includegraphics[width=\linewidth]{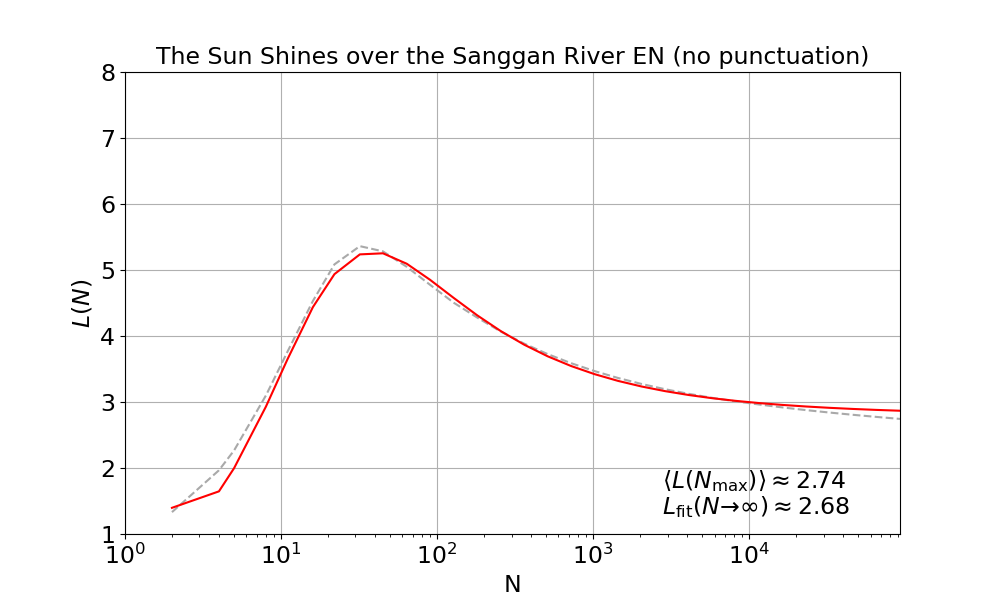}
\end{minipage}
\caption{Average shortest path length $L(N)$ as a function of network size $N$ for \textit{The Sun Shines over the Sanggan River} by Ding Ling (dashed) together with the fitted hybrid model (solid). Chinese original version with punctuation marks (a) and without punctuation marks (b) is shown together with its English translation with (c) and without punctuation marks (d). For each text, numerous growing network realizations were created by shifting the text starting point by a given number of words; the presented $L(N)$ was calculated as an average over all these realizations.}
\label{fig::sanggan_river}
\vspace{-0.4cm}
\end{figure*}

\begin{figure*}
\centering
\begin{minipage}[t]{0.48\linewidth}
\centering
(a)\includegraphics[width=\linewidth]{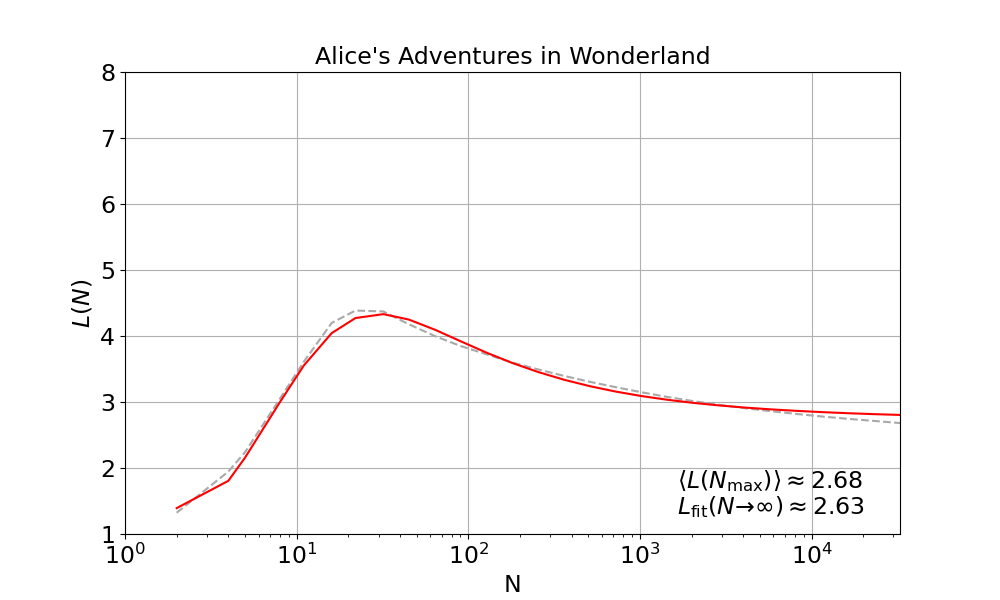}
\end{minipage}
\hfill
\begin{minipage}[t]{0.48\linewidth}
\centering
(b)\includegraphics[width=\linewidth]{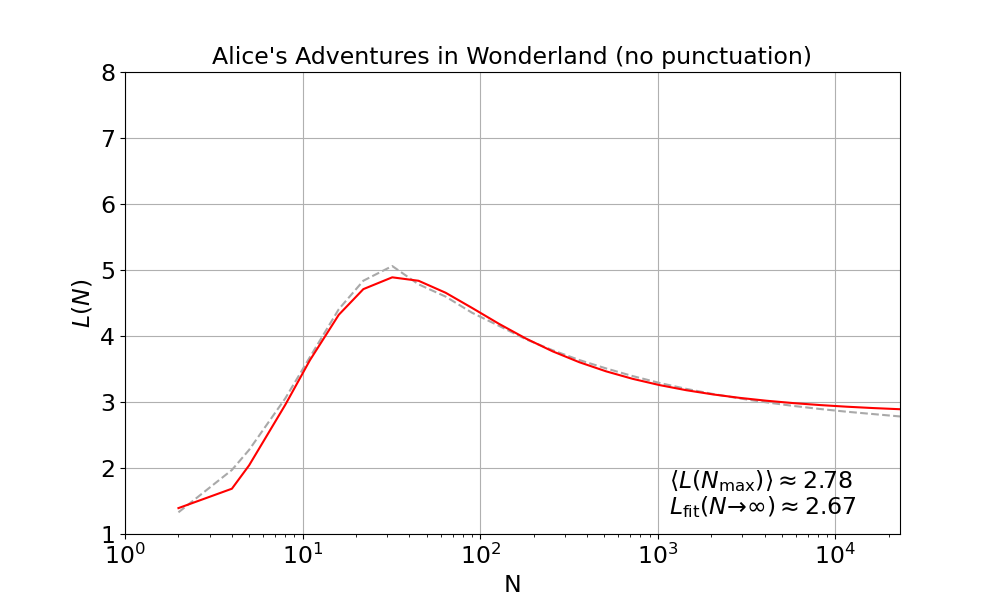}
\end{minipage}

\begin{minipage}[t]{0.48\linewidth}
\centering
(c)\includegraphics[width=\linewidth]{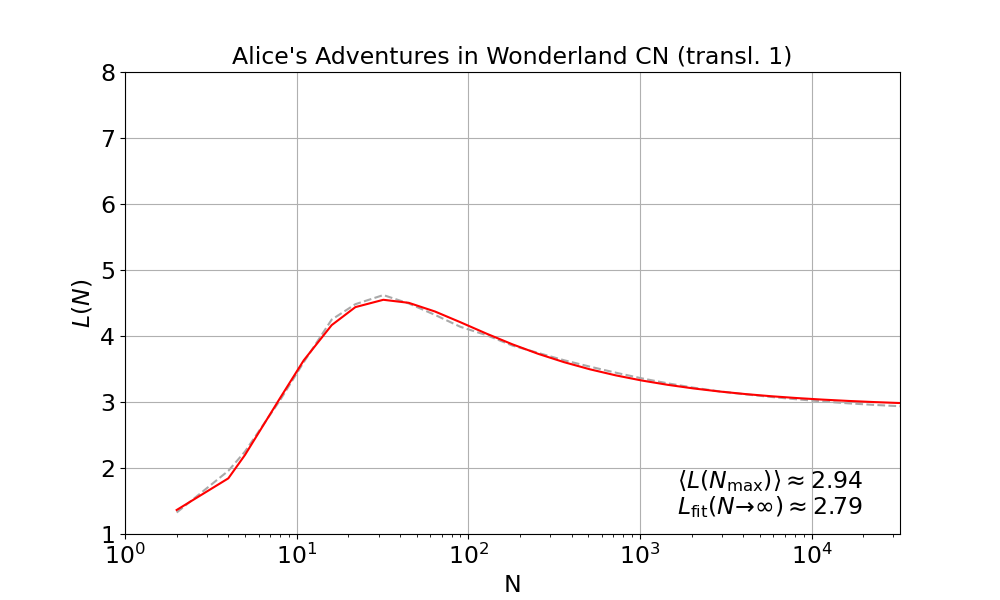}
\end{minipage}
\hfill
\begin{minipage}[t]{0.48\linewidth}
\centering
(d)\includegraphics[width=\linewidth]{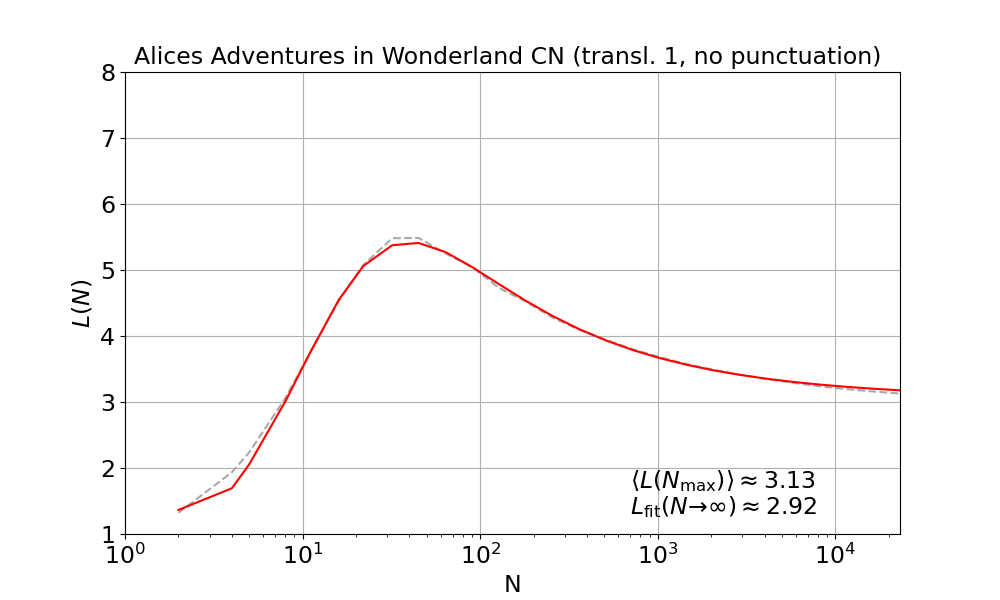}
\end{minipage}

\begin{minipage}[t]{0.48\linewidth}
\centering
(e)\includegraphics[width=\linewidth]{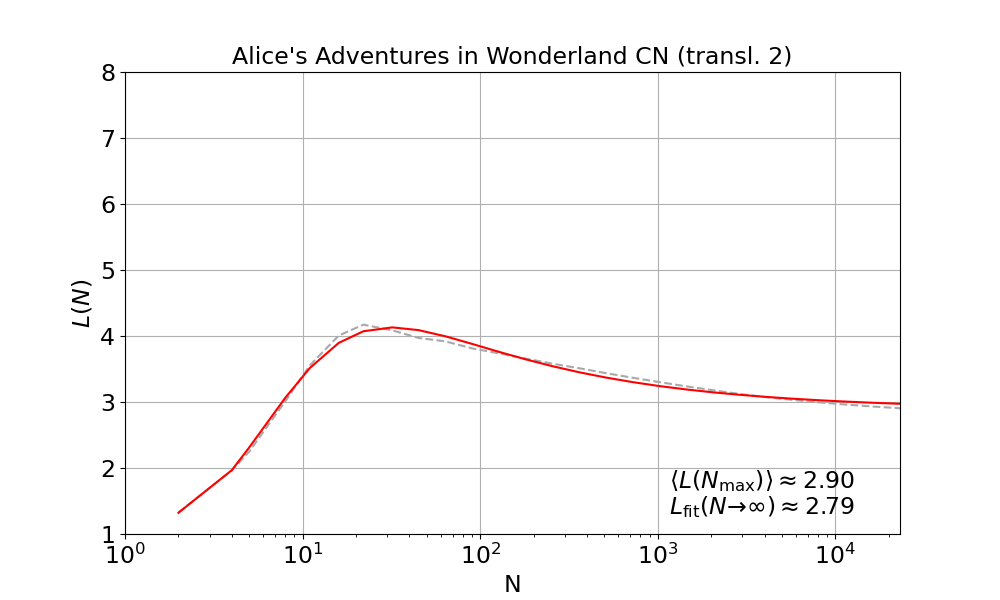}
\end{minipage}
\hfill
\begin{minipage}[t]{0.48\linewidth}
\centering
(f)\includegraphics[width=\linewidth]{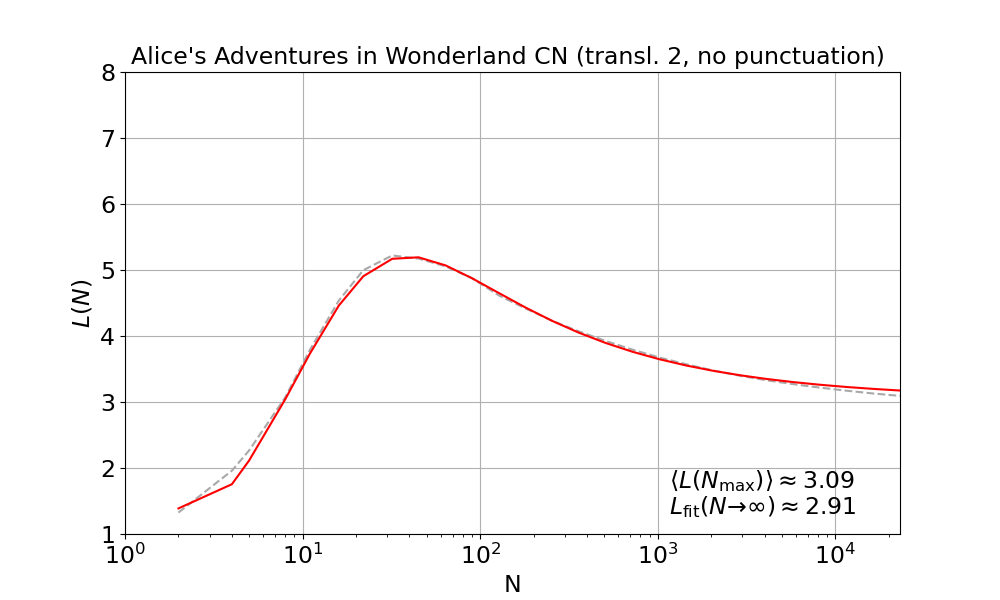}
\end{minipage}
\caption{Average shortest path length $L(N)$ as a function of network size $N$ for \textit{Alice's Adventures in Wonderland} by Lewis Carroll (dashed) together with the fitted hybrid model (solid). English original version with punctuation marks (a) and without punctuation marks (b) is shown together with its two Chinese translations with (c)(e) and without punctuation marks (d)(f). For each text, numerous growing network realizations were created by shifting the text starting point by a given number of words; the presented $L(N)$ was calculated as an average over all these realizations.}
\label{fig::alice_in_wonderland}
\vspace{-0.4cm}
\end{figure*}

\begin{figure*}
\centering
\begin{minipage}[t]{0.48\linewidth}
\centering
(a)\includegraphics[width=\linewidth]{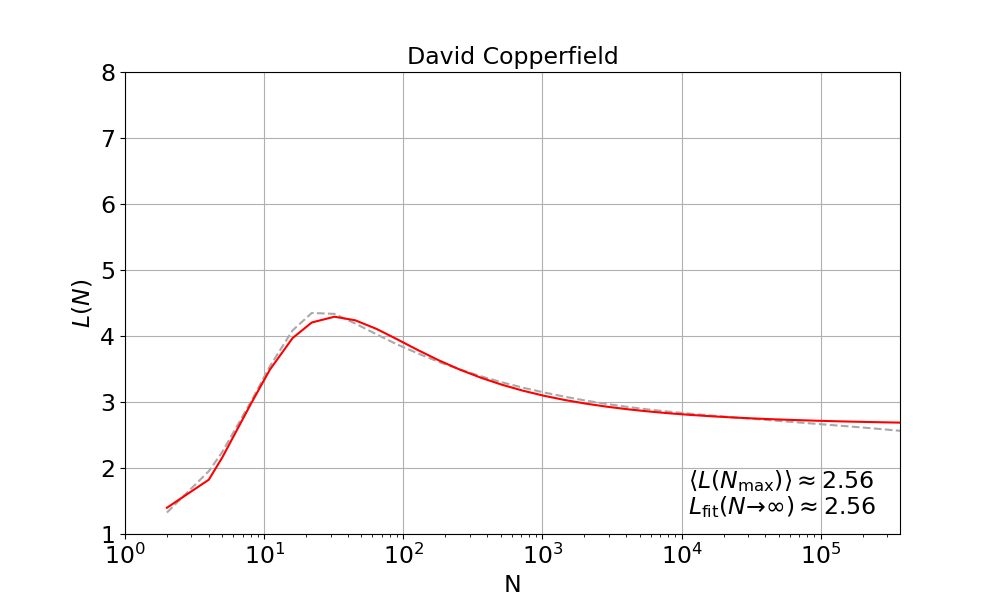}
\end{minipage}
\hfill
\begin{minipage}[t]{0.48\linewidth}
\centering
(b)\includegraphics[width=\linewidth]{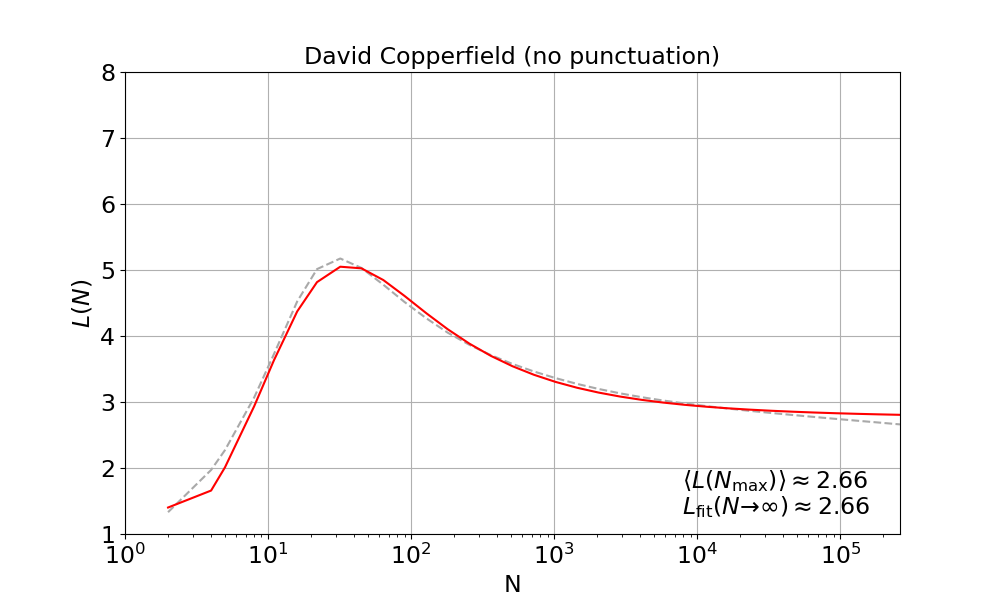}
\end{minipage}

\begin{minipage}[t]{0.48\linewidth}
\centering
(c)\includegraphics[width=\linewidth]{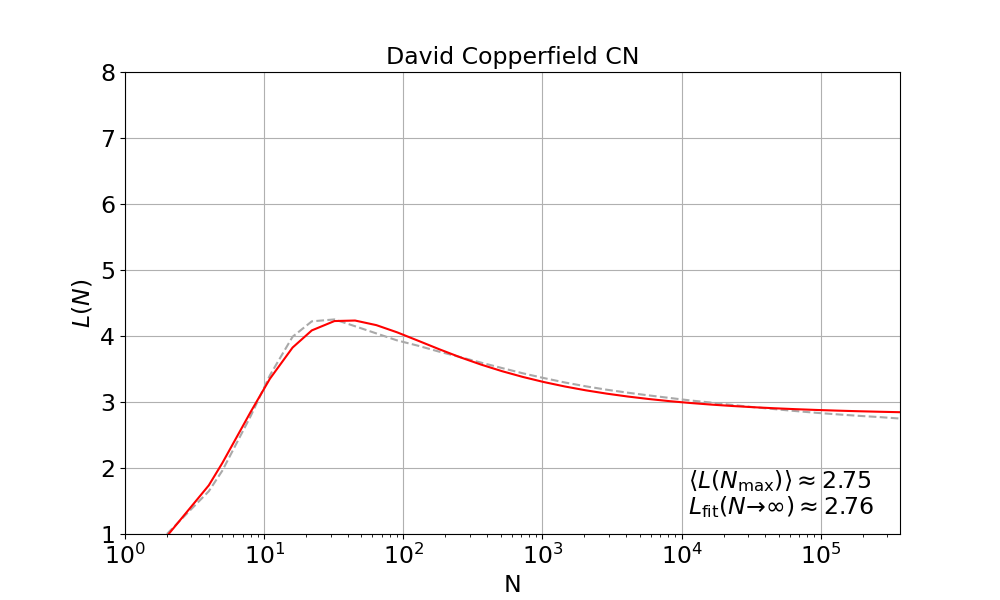}
\end{minipage}
\hfill
\begin{minipage}[t]{0.48\linewidth}
\centering
(d)\includegraphics[width=\linewidth]{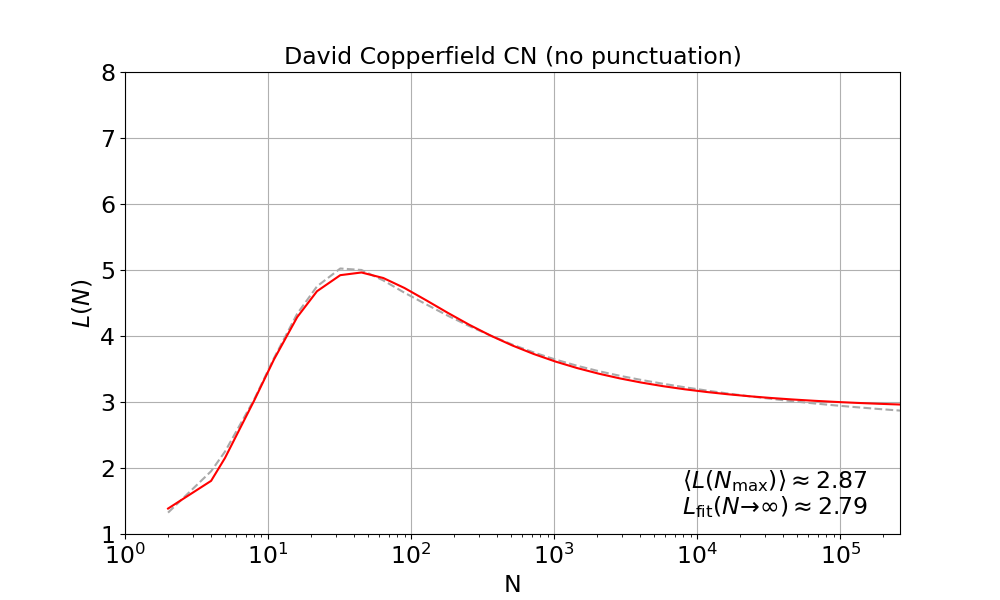}
\end{minipage}
\caption{Average shortest path length $L(N)$ as a function of network size $N$ for \textit{David Copperfield} by Charles Dickens (dashed) together with the fitted hybrid model (solid). English original version with punctuation marks (a) and without punctuation marks (b) is shown together with its Chinese translation with (c) and without punctuation marks (d). For each text, numerous growing network realizations were created by shifting the text starting point by a given number of words; the presented $L(N)$ was calculated as an average over all these realizations.}
\label{fig::david_copperfield}
\end{figure*}

To conduct a cross-linguistic analysis between Chinese and English, we will consider selected books originally written in Chinese that have been translated into English (denoted as C1-C3 in Sect.~\ref{sect::data}), as well as those originally written in English and translated into Chinese (denoted as E1-E2). Figs.~\ref{fig::drunkard}--\ref{fig::david_copperfield} show ASPL calculated for both types of network representations of each text, along with the fitted model $L_{\rm fit}(N)$. The relations between the properties of networks with and without punctuation are, in each case, similar to what was observed in Figs.~\ref{fig::chinese.corpora.1}--\ref{fig::chinese.corpora.2}. This is not surprising as each of the original Chinese texts (C1-C3) was already included in the appropriate collection of texts considered before. What is worth noting here is the relatively small impact of the CN $\to$ EN translation on the behavior of ASPLs, despite the significant differences between the grammatical and lexical structures of the two languages. This is especially true for the asymptotic behavior and limiting values $L(N \to \infty)$. An analogous translation invariance between various Western languages was observed earlier in the fractal properties of sentence-length variability and punctuation-mark distances~\cite{StaniszT-2023a,BartnickiK-2025a}. Slightly larger differences can be seen in the maximum values $L_{\rm max}$ for $10 < N < 100$, for both networks with punctuation and those without. These differences generally do not exceed approximately 0.5. The behavior of $L(N)$ and $L_{\rm fit}(N)$ for the complete texts with punctuation ($N = N_{\rm tot}$) and the asymptotic values $L(N \to \infty)$ are remarkably similar in the case of C1 and C2 ($\Delta L$ < 0.05, Figs.~\ref{fig::drunkard} and~\ref{fig::soul_mountain}), but differ noticeably for C3, E1, and E2 ($0.15 < \Delta L < 0.4$, Figs.~\ref{fig::sanggan_river}--\ref{fig::david_copperfield}). When punctuation marks are removed, the differences between the original texts become noticeable in all cases: $0.15 \leqslant \Delta L \leqslant 0.3$. An interesting situation can be seen in the two Chinese translations of \textit{Alice's Adventures in Wonderland}, whose network representations exhibit nearly identical asymptotic ASPL behavior (Fig.~\ref{fig::alice_in_wonderland}).

\begin{table*}
\caption{Frequency of the most common punctuation marks in Chinese texts considered in this work (C1-C3) and their English translations.}
\begin{ruledtabular}
\begin{tabular}{||c||c|c|c||c|c|c||c|c|c||}
 & \multicolumn{3}{c||}{C1: \textit{The Drunkard}} & \multicolumn{3}{c||}{C2: \textit{The Soul Mountain}} & \multicolumn{3}{c||}{C3: \textit{The Sun Shines over...}} \\
\hline
mark & CN (orig.) & EN (transl.) & ratio & CN (orig.) & EN (transl.) & ratio & CN (orig.) & EN (transl.) & ratio \\
 \hline\hline
. & 3,922 & 6,783 & 0.58 & 6,209 & 9,353 & 0.66 & 4,555 & 6,005 & 0.76 \\
\hline
, & 11,945 & 6,105 & 1.96 & 15,842 & 7,977 & 1.99 & 4,975 & 3,089 & 1.61 \\
\hline
! & 1,003 & 926 & 1.08 & 392 & 389 & 1.01 & 148 & 139 & 1.07 \\
\hline
? & 688 & 730 & 0.94 & 1,290 & 1,101 & 1.17 & 539 & 546 & 0.99
\end{tabular}
\end{ruledtabular}
\label{tab::punctuation.marks}
\end{table*}

\begin{figure*}
\centering
\begin{minipage}[t]{0.48\linewidth}
\centering
(a)\includegraphics[width=\linewidth]{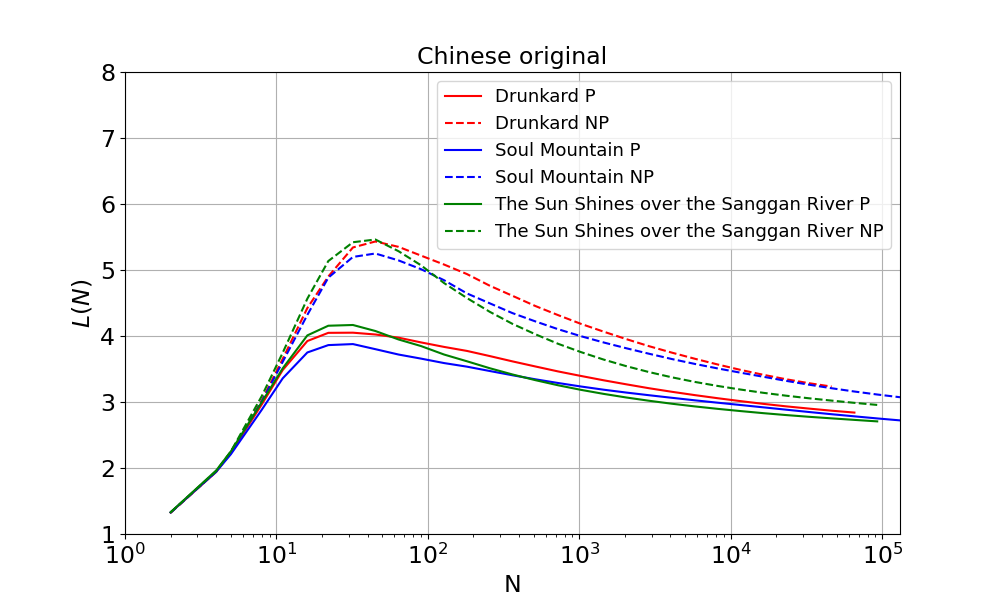}
\end{minipage}
\hfill
\begin{minipage}[t]{0.48\linewidth}
\centering
(b)\includegraphics[width=\linewidth]{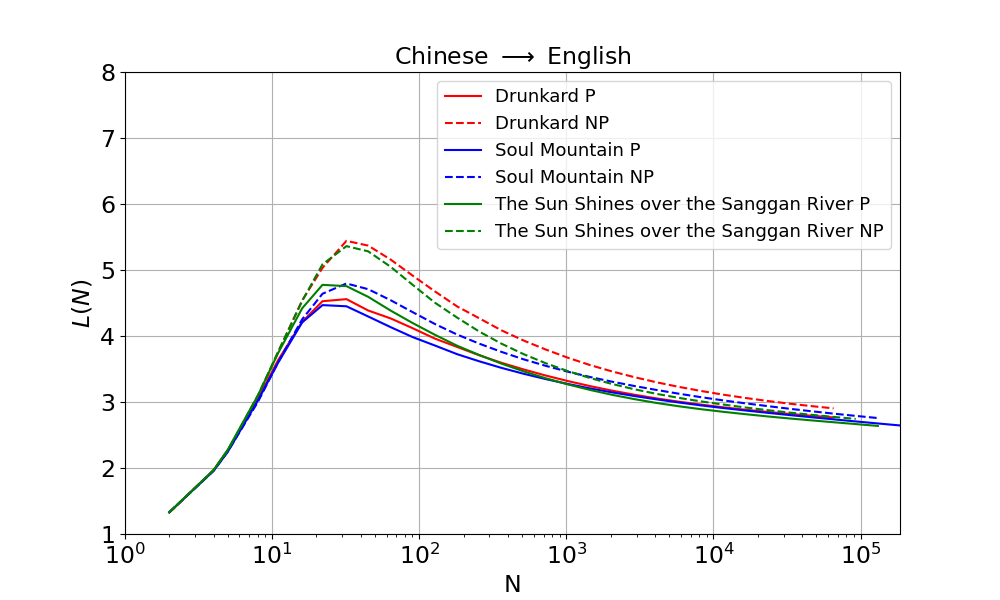}
\end{minipage}

\begin{minipage}[t]{0.48\linewidth}
\centering
(c)\includegraphics[width=\linewidth]{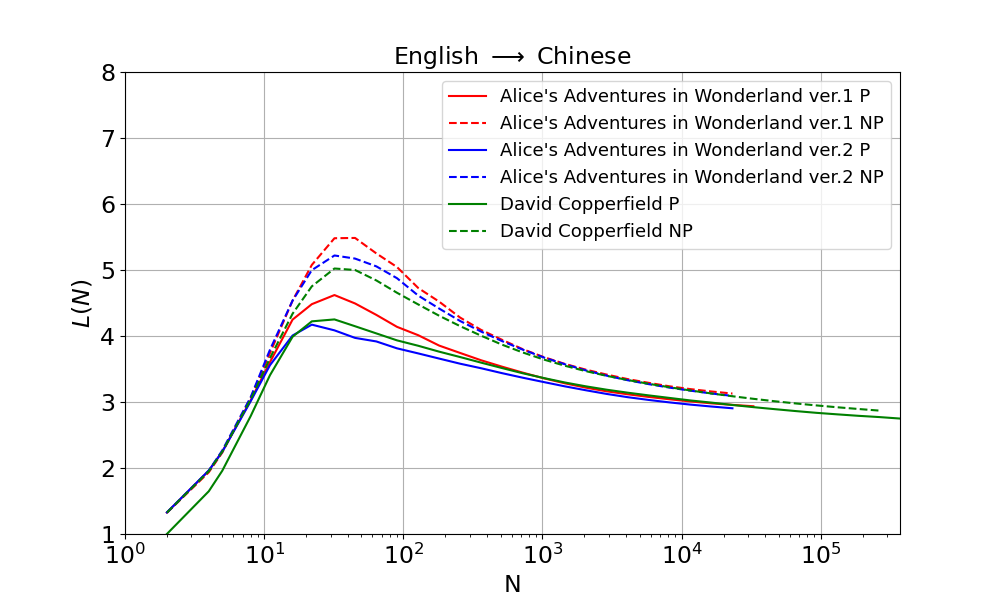}
\end{minipage}
\hfill
\begin{minipage}[t]{0.48\linewidth}
\centering
(d)\includegraphics[width=\linewidth]{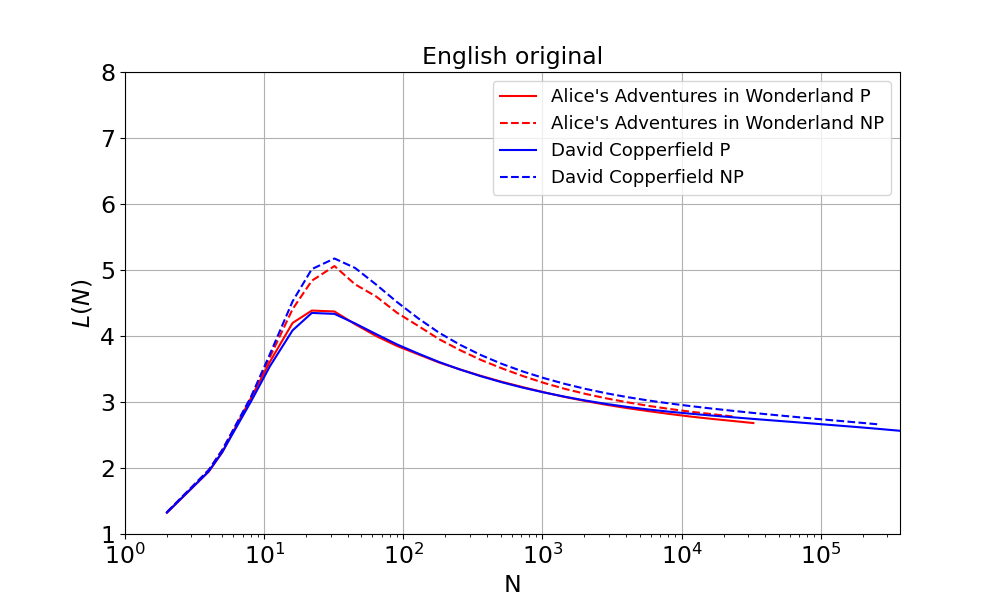}
\end{minipage}
\caption{Average shortest path length $L(N)$ as a function of network size $N$ for Chinese books translated to English (top) and English books translated to Chinese (bottom). Version with punctuation marks (solid) and without punctuation marks (dashed) are shown together for each book. For each text, numerous growing network realizations were created by shifting the text starting point by a given number of words; the presented $L(N)$ was calculated as an average over all these realizations.}
\label{fig::translation.matrix}
\end{figure*}

The dependence of ASPL on $N$ for the original texts and their translations, CN $\to$ EN and EN $\to$ CN, can be presented in a more compact form as a “translation matrix” -- see Fig.~\ref{fig::translation.matrix}. This arrangement allows one for an easy comparison of ASPL behavior for both network representations within each of the four text groups. The most prominent qualitative difference is the increase in $L(N)$ after punctuation marks are neglected, which is manifest in every case, both for networks in their infantile stage, where $N$ is small, and for large, mature networks, when asymptotic behavior is considered. It can also be observed that Chinese-language networks experience a stronger reduction in centralization after punctuation is removed than their English-language counterparts do. This applies to both the original texts and their translations. This suggests that the effect is a property of the languages themselves, regardless of whether a text was originally written in a given language or merely translated into it (with the caveat that the bilingual text set considered here is limited, and therefore this conclusion may not hold in general). Indeed, this property of Chinese comes from the fact that certain punctuation marks are more frequent in this language than they are in English: comma is the best example, which is much more frequently used in Chinese texts than in English ones -- Tab.~\ref{tab::punctuation.marks} documents it quantitatively.

\section{Conclusions}

We analyzed word-adjacency networks constructed from words and punctuation marks, treating both equally as lexical units carrying meaningful information. The main dataset consisted of texts written in Chinese, spanning a long period from the late 19th century to the present day. These texts were grouped by literary period, as well as by origin -- including texts by authors from Taiwan and Hong Kong, and texts published exclusively online. The network topology was examined through the node degree distribution and the average shortest path length. The networks exhibited a hierarchical character with a scale-free degree distribution characterized by an exponent of $\alpha \approx 2$. This value sets the studied networks apart from the Barabási-Albert networks, which follow linear, time-independent preferential attachment, and instead places them in the category of networks with accelerated growth and nonlinear degree-dependent attachment probabilities~\cite{KuligA-2015a}.

ASPL in Chinese texts demonstrates a similar dependence on the number of nodes as seen in Indo-European languages: for small $N$, it can be approximated by a linear function, and in the large-$N$ limit, by a monotonically decreasing function expressed by Eq.~(\ref{eq::aspl.large.network}). The interpolation between these two extremes provided a good fit. As intuitively expected, ASPL increased when only regular words were considered and punctuation marks were ignored. The largest differences were observed in the oldest group of texts, dating from the Late Qing period -- before Western-style punctuation was introduced into Chinese. Differences between the remaining groups, as measured by mean ASPL, were not substantial.

In addition to Chinese texts, the study also analyzed English translations, as well as Chinese translations of originally English texts. The shape of the mean ASPL functions was similar in both languages, regardless of whether token-adjacency networks or word-adjacency networks were considered. However, the asymptotic value of the fitted ASPL model was higher for Chinese than for English. This aligns with the observation that hubs with maximum degree in Chinese have slightly fewer connections to other nodes compared to their English counterparts, which results in a longer ASPL. For the token-adjacency networks, the differences were small to moderate, depending on the particular texts, whereas in word-adjacency networks, the differences increased and were substantial for all texts and both translation directions. Interestingly, omitting punctuation marks in the analysis caused a greater increase in ASPL for Chinese texts than for English ones. Our study has shown that translations between English and Chinese lead to smaller structural changes in the topology of word- and token-adjacency networks than it was observed earlier, for example, in a study of word-frequency and word-length time series in English and Esperanto texts by using fractal analysis~\cite{AusloosM-2012a}.

The observed decrease in the average shortest path length (ASPL) upon inclusion of punctuation indicates that punctuation acts as a set of topological shortcuts linking otherwise distant parts of the linguistic network. These marks reduce the average distance between words, thereby enhancing the network’s global connectivity and efficiency. In linguistic terms, punctuation contributes to text coherence by organizing syntactic and semantic boundaries — a function particularly important in Chinese, where word segmentation and grammatical marking are less explicit than in English. The fact that the ASPL reduction is more pronounced in Chinese implies that punctuation plays a relatively stronger structural role in this language, compensating for its morphosyllabic nature and absence of inflectional markers. In English, the effect is weaker because syntactic relations are more directly encoded in morphology and word order. 

For text classification, this means that punctuation not only contributes to stylistic and syntactic differentiation but also homogenizes network topology across texts. When punctuation is included, Chinese and English texts exhibit more similar small-world characteristics, suggesting that punctuation carries structural information useful for cross-linguistic stylometric or authorship-attribution tasks.

We focused on ASPL because it captures the global navigability and information accessibility of the language network, reflecting how efficiently different parts of a text can be reached through adjacency relations. While other network measures (e.g., clustering coefficient, modularity, betweenness) could complement the analysis, ASPL provides the most direct quantification of punctuation’s global integrative role across languages. Still, a natural next step in this line of research would be to explore these other topological characteristics of the networks and investigate how they are affected by punctuation and translation.

Finally, in more general and formal terms, the relation between punctuation distribution and legibility is non-linear. When punctuation is absent, the network elongates and ASPL rises, indicating weak connectivity and lower readability. As punctuation frequency increases, ASPL decreases because punctuation marks form bridges between syntactic units and enhance network efficiency. However, if punctuation were used excessively, the few available marks would become overrepresented hubs, producing a star-like topology with extremely short paths but low linguistic coherence. Thus, although ASPL continues to decline, cognitive legibility would deteriorate. Natural punctuation distributions in both Chinese and English appear to occupy the middle ground, which likely is optimal in minimizing ASPL while maintaining linguistic structure — a balance that through some kind of self-organization supports fluent comprehension and efficient information transfer. Deriving conditions for such an optimal balance from more basic principles emerges an intellectual challenge for future research which, optimally, should be associated with parallel empirical studies of the type such as those that found~\cite{AntiqueiraL-2007,AmancioDR-2013} correspondence between ASPL for linguistic world adjacency networks and legibility/comprehension of texts. However, those studies did not take punctuation into account. Considering punctuation in the context of linguistic networks and in ASPL in particular is an innovative proposal which, as the results presented here show, turns out to be very valid and promising, and indicates the direction of research.

\end{document}